\documentclass[final,3p,times]{elsarticle}  

\makeatletter
\def\ps@pprintTitle{%
 \let\@oddhead\@empty
 \let\@evenhead\@empty
 \let\@oddfoot\@empty
 \let\@evenfoot\@empty
}
\makeatother

\usepackage{amssymb}
\usepackage{amsmath}

\usepackage{cancel}
\usepackage{dashrule}
\usepackage[pdftex, pdfborderstyle={/S/U/W 1}]{hyperref}

\usepackage{xparse} 

\usepackage{xcolor}
\usepackage{algorithm}
\usepackage{algpseudocode}

\newcommand{\bs}[1]{{\boldsymbol{#1}}}
\newcommand{\bft}[1]{{\mathbf{#1}}}
\newcommand{\bb}[1]{{\mathbb{#1}}}

\newcommand{\bx}{\bft{x}}

\newcommand{\bz}{\bft{z}}

\newcommand{\bW}{\bft{W}}

\newcommand{\bsigma}{\bs{\sigma}}
\newcommand{\bmu}{\bs{\mu}}
\newcommand{\bSigma}{\bs{\Sigma}}
\newcommand{\btheta}{\bs{\theta}}

\newcommand{\bnabla}{\bs{\nabla}}
\newcommand{\bdiv}[1]{\bs{\nabla}{#1}\cdot}

\newcommand{\parenthesis}[1]{\left({#1}\right)}
\newcommand{\brackets}[1]{\left[{#1}\right]}

\newcommand{\llnorm}[1]{\left\lVert{#1}\right\lVert}

\NewDocumentCommand{\repeatindex}{>{\SplitList{,}}m}{%
  _{\begin{aligned}
    \ProcessList{#1}{\repeatindexhelper}%
  \end{aligned}}%
}
\newcommand{\repeatindexhelper}[1]{%
  & \scriptstyle #1 \\[-.7em]
}

\newcommand{\norm}[1]{\left\Vert{#1}\right\Vert}
\newcommand{\abs}[1]{\lvert{#1}\rvert}

\newcommand{\eref}[1]{Eq.~(\ref{#1})}

\newcommand{\demi}{\frac{1}{2}}
\newcommand{\id}{\mathrm{d}}
\newcommand{\iexp}{\mathrm{e}}
\newcommand{\itr}{\mathsf{T}}
\newcommand{\trace}{\text{Tr}}
\newcommand{\idata}{\text{data}}
\newcommand{\imin}{\text{min}}
\newcommand{\imax}{\text{max}}

\newcommand{\Rset}{{\mathbb R}}
\newcommand{\Cset}{{\mathbb C}}
\newcommand{\dataset}{{\mathcal D}}
\newcommand{\esp}{{\mathbb E}}
\newcommand{\Normal}{{\mathcal N}}
\newcommand{\Uniform}{{\mathcal U}}
\newcommand{\Loss}{{\mathcal L}}
\newcommand{\score}{\bft{s}}

\newcommand{\uv}{\bft{u}}

\newcommand{\Uv}{\bft{U}}
\newcommand{\rv}{\bft{r}}
\newcommand{\xv}{\bft{x}}

\newcommand{\Pres}{P}
\newcommand{\stress}{\bft{S}}
\newcommand{\density}{\varrho}
\newcommand{\Reynolds}{\text{Re}}
\newcommand{\Mach}{M}
\newcommand{\nablav}{\bs{\nabla}}
\newcommand{\zerov}{\bft{0}}
\newcommand{\Wiener}{\bft{W}}
\newcommand{\Iv}{\bft{I}}
\newcommand{\pdf}{p}
\newcommand{\qdf}{q}

\newcommand{\IC}{{\mathcal I}}
\newcommand{\BC}{{\mathcal B}}

\newcommand{\Tphysics}{{\mathcal T}}

\newcommand{\ELBO}{\operatorname{ELBO}}
\newcommand{\DKL}{\bb{D}_\text{KL}}
\newcommand{\MSE}{\text{MSE}}
\newcommand{\pearson}{\text{PCC}}
\newcommand{\denoiser}{\bs{\epsilon}}
\newcommand{\bphi}{\bs{\phi}}
\newcommand{\heat}{\bft{q}}
\newcommand{\ndim}{n}
\newcommand{\nsample}{S}
\newcommand{\Energy}{\mathcal{E}}
\newcommand{\Corr}{\bft{R}}
\newcommand{\Var}{\bb{V}}
\newcommand{\FFT}[1]{\widehat{#1}}
\newcommand{\unit}[1]{\hat{#1}}
\newcommand{\cjg}[1]{{#1}^*}
\newcommand{\taverage}[1]{\overline{#1}}
\newcommand{\saverage}[1]{\left\langle#1\right\rangle}
\newcommand{\kj}{\kappa}
\newcommand{\kv}{\bs{\kj}}
\newcommand{\fv}{\bft{f}}

\newcommand{\vortj}{\omega}
\newcommand{\vorticity}{\bs{\vortj}}
\newcommand{\ci}{\mathrm{i}}
\newcommand{\curl}{\bft{curl}}
\newcommand{\Tint}{\Delta\tau}
\newcommand{\TLya}{\tau_\text{L}}

\usepackage{color}

\newcommand{\sref}[1]{Sect.~\ref{#1}}
\newcommand{\tref}[1]{Table~\ref{#1}}
\newcommand{\fref}[1]{Figure~\ref{#1}}
\newcommand{\aref}[1]{Algorithm~\ref{#1}}

\begin{document}
\begin{frontmatter}



\title{Autoregressive regularized score-based diffusion models for multi-scenarios fluid flow prediction}

\author[label1,label2]{Wilfried Genuist}
\author[label1]{\'Eric Savin}
\author[label2]{Filippo Gatti}
\author[label2]{Didier Clouteau}

\affiliation[label1]{organization={DTIS, ONERA, Université Paris-Saclay},
            city={Palaiseau},
            postcode={91120}, 
            country={France}}

\affiliation[label2]{organization={LMPS-Laboratoire de Mécanique Paris-Saclay, Université Paris-Saclay, ENS Paris-Saclay, CentraleSupélec, CNRS},
            city={Gif-sur-Yvette},
            postcode={91190}, 
            country={France}}

\begin{abstract}

Building on recent advances in scientific machine learning and generative modeling for computational fluid dynamics, we propose a conditional score-based diffusion model designed for multi-scenarios fluid flow prediction. Our model integrates an energy constraint rooted in the statistical properties of turbulent flows, improving prediction quality with minimal training, while enabling efficient sampling at low cost. The method features a simple and general architecture that requires no problem-specific design, supports plug-and-play enhancements, and enables fast and flexible solution generation. It also demonstrates an efficient conditioning mechanism that simplifies training across different scenarios without demanding a redesign of existing models. We further explore various stochastic differential equation formulations to demonstrate how thoughtful design choices enhance performance. We validate the proposed methodology through extensive experiments on complex fluid dynamics datasets encompassing a variety of flow regimes and configurations. Results demonstrate that our model consistently achieves stable, robust, and physically faithful predictions, even under challenging turbulent conditions. With properly tuned parameters, it achieves accurate results across multiple scenarios while preserving key physical and statistical properties. We present a comprehensive analysis of stochastic differential equation impact and discuss our approach across diverse fluid mechanics tasks.

\end{abstract}




\end{frontmatter}




\section{Introduction}
\label{introduction}

With the rise of computational power and advanced numerical methods, simulating flow fields has become critical across many scientific and engineering domains. Among the most prominent frameworks for this purpose, Computational Fluid Dynamics (CFD) offers numerical tools to simulate and analyze fluid behavior governed by partial differential equations (PDEs), enabling accurate prediction of complex flow phenomena \cite{HIR07,POP00}. As a major discipline within fluid mechanics and computational physics, CFD drives innovation across energy production, transportation, environmental modeling, and industrial process optimization. It offers critical tools for simulating complex fluid behaviors in contexts ranging from aerodynamic design to renewable energy systems and pollutant dispersion. However, traditional solvers, despite their ability to produce high-fidelity results, often face difficulties in balancing computational cost and precision, particularly when addressing the multiscale and unpredictable nature of turbulent flows. These flows exhibit chaotic, highly nonlinear behavior with both long and short-lived spatiotemporal dynamics, making them especially difficult to model accurately. High-fidelity results can be obtained using numerical solvers based on finite difference, finite volume, or spectral methods, including Direct Numerical Simulation (DNS) \cite{DNS_base} and Large Eddy Simulation (LES) \cite{SCH22}. Yet, these physics-based, eddy-resolving models typically require large computational resources and are difficult to maintain, often restricting their application to simple designs and model evaluations.

Over the years, the evolution of machine learning (ML) and deep learning has significantly transformed the field of CFD. These rapid advances have introduced innovative approaches to the field, including deep learning-based closure models or discretization techniques; see for example \cite{brunton2020machine,BUZ23,DUR19,LIN23,VIN22}. A leading example of ML approaches for solving PDEs is physics-informed neural networks \cite{raissi2019physics} (PINNs, with early developments in \cite{LAG98,LAG00}), which integrate the governing equations directly into the training process. While continuously improved, these models can suffer from unstable training and high data requirements. Their effectiveness significantly declines when applied to turbulent flows due to their chaotic nature, which conflicts with the deterministic assumptions underlying most traditional ML models. Other advanced deep learning architectures, including graph neural networks \cite{GNN09}, Fourier neural operators (FNOs) \cite{LI21FNO,wang2024prediction}, and operator learning algorithms such as DeepONets \cite{CHE95,LU21}, have shown strong performance on a variety of fluid dynamics problems. However, achieving long-term stability and accuracy remains a tremendous challenge. In parallel, foundation models such as Poseidon \cite{herde2024poseidon}, designed to learn solution operators of PDEs, have emerged as versatile and efficient frameworks for general PDE resolution, while PDE-Refiners have specifically focused on stable and accurate long-term predictions \cite{lippe2023pde}. More recently, Kolmogorov–Arnold Networks (KANs) \cite{liu2025KAN} have been proposed, with applications extending to fluid dynamics and statistical forecasting \cite{kanFluid}.

On the other hand, generative modeling has enabled the simulation of complex flows with promising accuracy. Its probabilistic and fast sampling nature makes it well-suited for uncertain or out-of-distribution scenarios, providing a coherent and efficient approach to turbulence modeling challenges. Generative Adversarial Networks (GANs) \cite{GAN2014} have been used to synthesize turbulent flows in \cite{DRY22} and compared with Variational Autoencoders (VAEs) \cite{Kingma2013} and Denoising Diffusion Probabilistic Models (DDPMs) \cite{Sohl2015deep,DDPM} in \cite{DRY24}. Additionally, general PDE solvers employing ML techniques have ventured into super-resolution tasks to enhance spatial resolution, as in \cite{fan2025neural} for example. The more complex task of three-dimensional (3D) turbulence prediction is addressed in \cite{lienen2023zero}, which proposes a zero-shot inference strategy for physically accurate 3D turbulence modeling. In the same spirit, diffusion-based generative models, particularly those grounded in flow and diffusion processes, offer a compelling solution for fast problem solving. By learning mappings between complex data distributions and simpler latent spaces (typically Gaussian), they enable efficient sample generation and likelihood estimation. Several generative approaches have been proposed to bridge the gap between data space and latent space, each offering distinct trade-offs. Normalizing flows \cite{normalizingFlows} use a sequence of invertible and differentiable transformations, enabling tractable likelihood estimation, but often require careful tuning. Flow matching \cite{flowMatching} reframes the task as learning a vector field that deterministically connects prior and data distributions via neural ordinary differential equations \cite{CHE18}, balancing efficiency and expressiveness with the challenge of training stability. Diffusion Schr\"odinger bridges \cite{schroBridge} leverage time-reversal formulations of stochastic processes to interpolate between distributions, offering expressive modeling of intermediate states. Similarly, stochastic interpolants \cite{stochInterpo} blend diffusion and optimal transport principles to design semi-deterministic transition paths. 

Lately, attention has shifted towards diffusion models for addressing PDE-related problems, valued for their flexibility and robustness \cite{shysheya2024conditional}. They have achieved significant results across a variety of tasks, from flow field reconstruction to super-resolution and inverse problems, as evidenced by benchmarking studies \cite{benchmarking}, partially observed systems as DiffusionPDE \cite{huang2024diffusionpde}, conditional latent models such as CoNFiLD \cite{du2024confild}, high-resolution applications \cite{fan2025neural} including weather forecasting \cite{GenCast25}, or turbulence scaling \cite{WHI24}. Notable developments include PDE-Refiner \citep{lippe2023pde} and unified time denoising approaches \cite{ruhling2023dyffusion}. Furthermore, physics-aware and physics-informed variants \cite{qiu2024pi, bastek2024physics, shu2023physics} have advanced interpretability and accuracy in complex fluid dynamics by incorporating the underlying PDE in the training process. Despite these advances, scaling remains a significant challenge, particularly for multi-field problems that are either insufficiently explored or require highly specialized architectures and considerable computational resources.

Addressing such issues requires careful design of conditioning mechanisms to effectively guide the solution. However, no definitive consensus has emerged regarding the optimal approach. Initial work by \citet{yang2023denoising} proposed incorporating desired constrains inside the model. Building on this, \citet{benchmarking} and \citet{shysheya2024conditional} highlighted the autoregressive formulation as a promising framework for multi-field forecasting with reduced architectural complexity by predicting future states based on past observations. Conditional latent diffusion methods, such as CoNFiLD \citep{du2024confild}, proposed an alternative approach by leveraging guided latent spaces for PDE solution generation. Meanwhile, the aforementioned PINN-like diffusion models by \citet{qiu2024pi} and \citet{bastek2024physics} have required additional conditioning models or adapted residual computation to obtain physical accuracy with generative flexibility. In parallel, conditional score-based generative models have gained attraction for addressing inverse problems in mechanics and integrating prior observations \citep{dasgupta2024conditional}. To overcome the ill-posedness and potential physical inconsistencies, other works have embedded auxiliary observation priors to guide generation toward physically plausible outcomes \citep{huang2024diffusionpde,gao2025generative}.  Similarly, \citet{jacobsen2025cocogen} introduced CoCoGen, a physically consistent, conditioned score-based framework for both forward and inverse problems. It incorporates additional post-training conditioning mechanisms inspired by text-to-image diffusion techniques \citep{zhang2023adding}. Other physics-informed score-based diffusion processes have expanded their applicability to physics-based inverse problems \citep{han2024physics}. This need for broader applicability is reflected in recent benchmarking efforts; for instance, \citet{shysheya2024conditional} present a comprehensive evaluation of conditional diffusion models for PDE simulation, analyzing various conditioning strategies and identifying persistent challenges, such as long-term prediction accuracy and time series consistency. In the broader context of guided generative modeling for parametric PDE solutions, the role of the underlying Stochastic Differential Equation (SDE) (which governs the generative process of score-based diffusion models) remains relatively underexplored. Most existing methods rely on standard SDE formulations without conducting thorough comparative analysis.

Overall, diffusion-based methods for fluid dynamics and time series solutions have mainly focused on DDPMs \cite{benchmarking,GenCast25} and, more recently, a unified framework with stochastic interpolants \cite{MUC25}. While these approaches demonstrate strong potential and flexibility, significant challenges remain. Despite the dominance of DDPMs, exploring score-based methods in fluid dynamics holds promises for achieving improvements with minimal adjustments. Generalizing from 2D to 3D and achieving long-term predictions across diverse scenarios present significant challenges, highlighting the need for robust regularization strategies. Key areas for progress include optimized noise schedules in SDE-based methods, improved conditioning, and more systematic application to fluid problems. Current models often favor complex architectures over simpler, well-tuned alternatives. In this work, we show that a simple architecture, paired with efficient training, can yield statistically accurate and long-term stable predictions, while maintaining fast training and sampling across a range of 2D and 3D cases, without requiring model redesign. To the best of our knowledge, this is the first conditional score-based diffusion model capable of full-time fluid flow prediction using multiple SDE types across multi-field and multi-scenario settings, balancing performance and lightweight design.

The remainder of the paper is structured as follows. In \sref{sec:discrete_setting}, we provide an overview of score-based diffusion models in both discrete-time and continuous-time settings. We then focus on the autoregressive formulation and the conditional score-matching objective used to train the inference process. Datasets and implementation of the proposed autoregressive, score-based diffusion models are then outlined in \sref{sec:data_and_archi}. A particular attention is given to the selection of SDE models and the development of a regularization procedure aimed at improving the stability of inferences for fluid flow reconstruction. The results are presented in \sref{sec:Results} for the three datasets considered in previous works, namely, a compressible transonic flow about a cylinder in two dimensions \cite{benchmarking}, a turbulent radiative layer between hot and cold gases in two dimensions \cite{2DradLay}, and a three-dimensional flow coupled with magnetodynamics \cite{MHD64}. Comparisons of the predictions of diffusion models with the ground-truth solutions are performed with dedicated metrics relevant to CFD. The analysis highlights how well the models capture the complex, chaotic structures characteristic of turbulent flows. A study is conducted to evaluate the effects of a specific filtering technique which proves to be beneficial in reducing residual noise without compromising key flow features. This synergy between filtering and regularization enhances both the stability and accuracy of the predictions. Some conclusions and perspectives are finally drawn in \sref{sec:conclusion}.

\section{Foundations of score-based diffusion models}
\label{sec:discrete_setting}

In this section the score-based, generative diffusion models considered throughout the paper are outlined. We start by introducing the discrete-time setting corresponding to DDPM \cite{Sohl2015deep,DDPM} in \sref{backgroundDDPM}, before we turn to the continuous-time setting corresponding to the interpretation of the generative process in terms of forward and backward SDE \cite{SMSDE} in \sref{continuous setting}. The autoregressive formulation adapted to fluid flow reconstruction, and the associated conditional score-matching objective are then detailed in \sref{sec:score-matching}.

\subsection{Background on generative models and denoising diffusion probabilistic models}
\label{backgroundDDPM}

Generative models aim to sample from an unknown probability distribution $\pdf_\idata$ that describes some dataset $\dataset$ assuming that $\bx_{\text{sample}} \sim \pdf_\idata(\bx)$. Here the frame $\bx\in\Rset^\ndim$ where $\ndim$ is typically a large integer (\emph{e.g.} the number of pixels in image generation), and $\dataset=\{\bx^{(s)}\}_{s=1}^\nsample$ contains $\nsample$ sample frames from $\pdf_\idata$. Within a machine learning framework, the objective is to construct an appropriate probability density function $\pdf_\btheta$ (in energy models, a flexible energy function that is then normalized) with parameters $\btheta$ to model this density from its samples in $\dataset$. For the sake of simplicity, such samples are considered as independent and identically distributed. Under a given metrics (for example, the Kullback–Leibler divergence $\DKL$), the likelihood is then maximized over all samples in the dataset (likelihood-based approach), seeking optimal parameters $\btheta^*\in \arg \min_\btheta \DKL(\pdf_\idata\|\pdf_\btheta)$ where:
\begin{equation}
\DKL(\pdf\|\qdf) := \esp_{\bx\sim\pdf(\bx)}\{ \log\pdf(\bx) - \log\qdf(\bx)\}\,.
\end{equation}
Here and throughout the paper $\esp$ stands for the mathematical expectation, or the ensemble average. Samples can be generated by various methods, such as direct sampling or Markov Chain Monte-Carlo (MCMC) methods \cite{AND03}, once $\pdf_{\btheta^*}$ is known. Alternatively with VAEs \cite{Kingma2013}, one sets a latent space $\bz \sim \pdf(\bz)$ and learn forward $\qdf_\bphi(\bz \lvert \bx)$ and backward $\pdf_\btheta(\bx \lvert \bz)$ mappings between the data and latent spaces. In this setup, new samples are obtained by drawing a sample $\bz$ in the latent space and applying the learned reverse transformation $\pdf_{\btheta^*}(\bx \lvert \bz)$ to generate new samples in the data space. The maximization objective reads:
\begin{equation}
\btheta^*,\bphi^*\in\arg\max_{\btheta,\bphi} \ELBO(\bx;\btheta,\bphi) \,,
\end{equation}
where $\ELBO$ stands for the Evidence Lower Bound:
\begin{equation*}
\begin{split}
\ELBO(\bx;\btheta,\bphi) &:= \esp_{\bz\sim\qdf_\bphi(\bz \lvert \bx)}\left\{\log{\frac{\pdf(\bz)\pdf_\btheta(\bx \lvert\bz)}{\qdf_\bphi(\bz \lvert \bx)}}\right\} \\
&= \esp_{\bz\sim\qdf_\bphi(\bz \lvert \bx)}\left\{\log\pdf_\btheta(\bx \lvert\bz)\right\} - \DKL(\qdf_\bphi(\cdot\lvert \bx)\|\pdf)\,.
\end{split}
\end{equation*}
The prior matching term $\DKL(\qdf_\bphi(\cdot\lvert \bx)\|\pdf)$ tells how good the encoder $\qdf_\bphi(\bz \lvert \bx)$ is versus a prior belief $\bz\sim\pdf(\bz)$ held over latent variables $\bz$, and has to be minimized to maximize the $\ELBO$. The reconstruction term $\esp_{\qdf_\bphi(\bz\lvert\bx)}\{\log\pdf_\btheta(\bx \lvert\bz)\}$ tells how good the decoder $\pdf_\btheta(\bx \lvert\bz)\approx\pdf_\idata(\bx\lvert\bz)$ is, and has to be maximized to maximize the $\ELBO$. Since $\log\pdf_\idata(\bx)\geq\ELBO(\bx;\btheta,\bphi)$, sampling $\bz\sim\pdf(\bz)$ and then $\bx\sim\pdf_{\btheta^*}(\bx \lvert \bz)$ after maximizing $\ELBO \parenthesis{\bx;\btheta,\bphi}$ leads to maximize the marginal $\pdf_\idata$ on average. 

Diffusion models \cite{Sohl2015deep,DDPM,SMSDE} decompose this process into $N-1$ intermediate states $\bx_i\sim\pdf_i(\bx_i)$, $1\leq i\leq N-1$ with $\bx_{0}\equiv\bx$ and $\bx_{N}\equiv\bz$, that establishes a mapping between the targeted data distribution $\pdf_0(\bx_0):=\pdf_\idata(\bx)$ and the latent distribution $\pdf_N(\bx_N):=\pdf(\bz)$. By specifying the form of the forward steps (successive Gaussian noise additions), the forward process slowly converges to a centered Gaussian distribution in a Markovian way, while ensuring dimension consistency between the states. In DDPMs \cite{Sohl2015deep,DDPM}, the forward process is given by:
\begin{equation}\label{eq:transition-1}
\qdf(\bx_{1},\dots\bx_{N} \lvert \bx_{0})
=\prod_{i=0}^{N-1}\qdf_{i+1 \lvert i}(\bx_{i+1} \lvert \bx_{i})
=\prod_{i=0}^{N-1}\Normal\left(\bx_{i+1} ; \sqrt{\alpha_{i+1}}\,\xv_i,(1-\alpha_{i+1})\Iv\right)\,,
\end{equation}
where $\Iv$ is the identity matrix, $\Normal$ is the usual Gaussian (normal) distribution, and $\{\alpha_i\}_{1\leq i \leq N}$ is a (usually linear) variance schedule \cite{DDPM} ensuring $\smash{\bx_N} \sim p(\bz)= \Normal(\bz;\zerov,\Iv)$ with enough steps $N$. The derivation of the maximization objective is similar to the one of the VAE and it can be shown \cite{DDPM,unified2022} that the $\ELBO$ objective is equivalent to the sum of denoising matching terms:
\begin{equation}\label{eq:denoising-matching}
   \btheta^* \in \arg\min_{\btheta} \sum_{i=2}^{N} \esp_{\qdf_{i \lvert 0}(\bx_{i} \lvert \bx_{0})}\left\{
\DKL(\qdf(\bx_{i-1} \lvert \bx_{i}, \bx_{0}), \pdf_\btheta(\bx_{i-1} \lvert \bx_{i}))
\right\}\,,
\end{equation}
where $\pdf_\btheta(\bx_{i-1} \lvert \bx_{i}) = \Normal\left(\bx_{i-1}; \bft{m}_\btheta(\bx_i, i), \bft{v}_\btheta(\bx_i,i)\right)$ seeks to reverse the addition of noise (thus also being Gaussian to imitate the forward step), forming the backward process of each state. After simplifications \cite{DDPM}, one may rewrite the problem \eqref{eq:denoising-matching} as of finding optimal parameters $\btheta^*$ of a denoiser model $\denoiser_\btheta(\bx_i,i)$:
\begin{equation}\label{eq:denoising}
\btheta^*\in\arg\min_\btheta
\esp_{
\begin{aligned}
& \scriptstyle \bx_0 \sim \pdf_0(\bx_{0}) \\[-.8em]
& \scriptstyle i \sim \Uniform(1,N) \\[-.8em]
& \scriptstyle \denoiser \sim \Normal(\zerov,\Iv)\\[-.8em]
\end{aligned}
}
\left\{ 
\left\lVert 
\denoiser - 
\denoiser_\btheta\parenthesis{\sqrt{\bar{\alpha}_i}\bx_{0}
+ \sqrt{1-\bar{\alpha}_i}\denoiser, i}
\right\lVert_{2}^{2}
\right\}\,, \quad \bar{\alpha}_i=\prod_{j=1}^i\alpha_j\,, 
\end{equation}
where $\Uniform(1,N)$ stands for the discrete uniform distribution over ${1,2,\dots,N}$ (see \ref{Nested expectations} for the notation of nested expectations). Also $\smash{\norm{\bft{a}}_2^2}=\smash{\sum_{j=1}^\ndim a_j^2}$ is the usual (squared) Euclidian norm of $\bft{a}\in\Rset^\ndim$. Simultaneously, the mean $\bft{m}_{\btheta^{*}}$ and the variance $\bft{v}_{\btheta^{*}}$ are retrieved by:
\begin{equation*}
\bft{m}_{\btheta^*}(\bx_i,i)=\frac{1}{\sqrt{\alpha_i}}\left(\bx_i-\frac{1-\alpha_i}{\sqrt{1-\bar{\alpha}_i}}\denoiser_{\btheta^*}(\bx_i, i)\right)\,,\quad \bft{v}_{\btheta^*}(\bx_i,i)= \frac{(1-\bar{\alpha}_{i-1})(1-\alpha_i)}{1-\bar{\alpha}_i}\Iv\,.
\end{equation*}
New samples can be drawn from $\pdf_0$ by generating pure Gaussian noise from the latent space $\bz\sim\pdf(\bz)$ and recursively denoising that latent variable over the backward steps. The denoiser model $\denoiser_\btheta$ is typically a deep neural network such as a U-Net \cite{UNet}, as discussed below in \sref{sec:UNet}.

\subsection{Score-based diffusion models}
\label{continuous setting}

\begin{figure}[t]
    \centering
    \includegraphics[width=\linewidth]{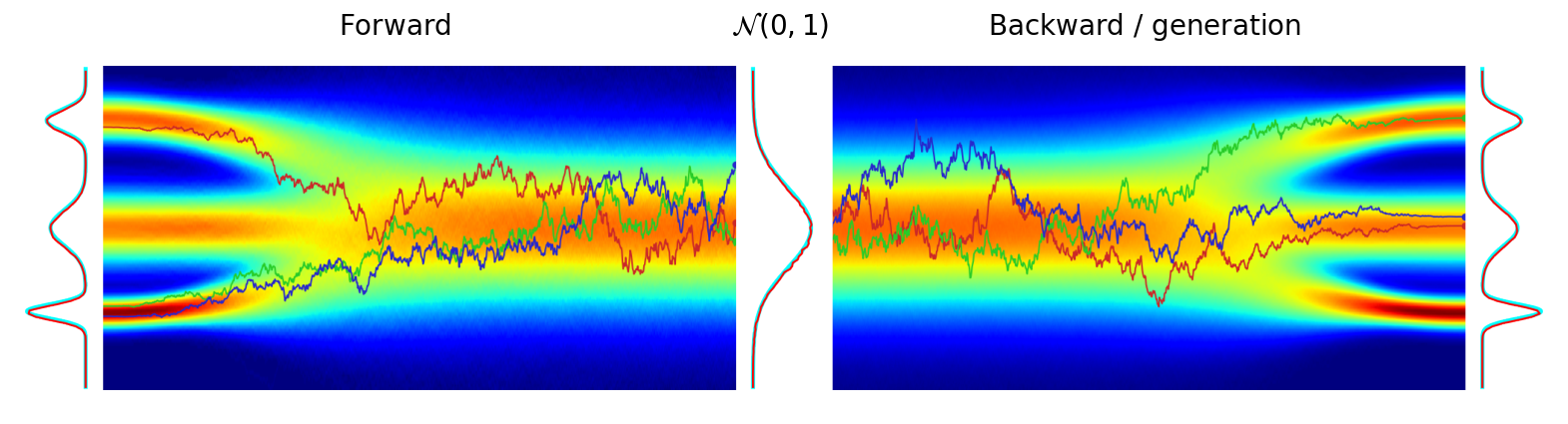}
    \vspace{-1.5em}
    \caption{Illustration of the link between the Fokker-Planck equation (FPE) and stochastic differential equation (SDE) trajectories. Left: The forward SDE evolution, starting from a Gaussian mixture, evolving towards a normal distribution $\mathcal{N}(0,1)$. Right: The backward (generation) process reconstructs the structured state by reversing the diffusion process. Colored lines represent individual stochastic trajectories governed by the underlying SDE, while the background color map depicts the evolving probability density as described by the FPE.}
    \label{fig:KPE_illus}
\end{figure}

Score-based diffusion models, as introduced by \citet{SMSDE}, are a class of generative diffusion models that generalize DDPM to a continuous-time framework. Their foundation lies in the non-normalized density estimation of energy-based models \cite{hyvarinen2005estimation,Song2019} achieved via the score function (also referred to as the Stein score function), defined as the gradient of the log probability density function:
$\score(\bx)=\nablav_\bx\log\pdf(\bx)$. The main difference from DDPM comes with the use of a continuous-time variable $\bx_t$ instead of discrete states $\bx_i$. The intuition behind the use of stochastic differential equations (SDE) in diffusion models is its link with the Fokker-Planck equation \cite{oksendal2013stochastic,sarkka2019applied}. Considering the expression of the transition kernel of the discrete setting in \sref{backgroundDDPM}, and changing the discrete states $\bx_i$ into continuously evolving states $\bx_t$ instead, the instantaneous change in time of the marginal distribution $\pdf_t(\bx_t)$ satisfies such a Fokker-Planck equation (see \fref{fig:KPE_illus}, with visualization inspired by \citet{SMSDE} and adapted from St\'ephane Mallat’s course\footnote{\href{https://github.com/jecampagne/cours_mallat_cdf/blob/main/cours2024/ScoreDiffusionGene.ipynb}{https://github.com/jecampagne/cours$\_$mallat$\_$cdf/blob/main/cours2024/ScoreDiffusionGene.ipynb}}).
Let $\{\bx_t;\,0\leq t<\infty\}$ be a random process defined in $\Rset^\ndim$ and indexed on $t\in\Rset_+$. It is a diffusion process if it is the solution of the It\={o} SDE \cite{oksendal2013stochastic} describing the evolution of a sample $\bx_0$ to its final state $\bx_T$ according to:
\begin{align}\label{SDE}
\id\bx_t=\bmu(\bx_t,t)\id t+\bsigma(\bx_t,t)\id\Wiener_t\,,\quad \bx_0\sim\pdf_0\,,
\end{align}
where $t\mapsto\bmu(\cdot,t):\Rset^\ndim \times \Rset \to \Rset^\ndim$ is the drift coefficient, $t\mapsto\bsigma(\cdot,t):\Rset^\ndim \times \Rset \to \Rset^{\ndim\times\ndim}$ is the diffusion coefficient, assumed to be continuous functions of time, and $\{\Wiener_t;t \geq 0\}$ is a $\ndim$-dimensional Brownian motion (or Wiener process). \citet{Anderson} showed that the time can be reversed $t\to T-t$ in $\eqref{SDE}$ which turns out to be also a diffusion process \cite{cattiaux2023time,haussmann1986time}. This yields the reverse-time, or backward SDE:
\begin{equation}\label{eq:BSDE}
\id\bx_t=\tilde{\bmu}(\bx_t,t)\id t + \bsigma(\bx_t,t)\id\overline{\Wiener}_t\,,\quad\bx_T\sim\pdf_T \,,
\end{equation}
with $\tilde{\bmu}(\bx,t)=\bmu(\bx,t)-\bnabla_\bx\cdot \bSigma(\bx,t)-\bSigma(\bx,t)\nablav_\bx\log \pdf_t(\bx)$, $\bSigma(\bx,t)=\bsigma(\bx,t)\bsigma(\bx,t)^\itr$ where $\bsigma^\itr$ is the transpose matrix of $\bsigma$, and $\{\overline{\Wiener}_{t}\}_{t\geq 0}$ is a time reversed Brownian motion. Its equivalent deterministic form (in the sense that their solutions have the same marginal distributions at each timestep $t$) is the Probability-Flow Ordinary Differential Equation (PF-ODE) \cite{SMSDE}:
\begin{equation}\label{eq:PF-ODE}
\frac{\id\bx_t}{\id t}=\bmu(\bx_t,t)-\demi \parenthesis{\bnabla_\bx\cdot \bSigma(\bx_t,t)- \bSigma(\bx_t,t) \bnabla_\bx\log \pdf_t(\bx_t)} \,,\quad\bx_T\sim\pdf_T \,.
\end{equation}
Similarly to the discrete-time approach in \sref{backgroundDDPM}, solving the backward SDE \eqref{eq:BSDE} or PF-ODE \eqref{eq:PF-ODE} starting from a sample $\xv_T\sim\pdf_T$ leads to the generation of a new sample from $\pdf_0$. This can be done by different methods, such as the usual Euler–Maruyama scheme for the backward SDE:
\begin{equation*}
\bx_{i} = \bx_{i+1} + \tilde{\bmu}(\bx_{i+1},i+1) \id t  + \bsigma_{i+1}\bz_{i+1}\!\sqrt{\id t}\,,\quad\bz_{i+1}\sim\Normal(\zerov,\Iv)\,,
\end{equation*}
or the forward Euler scheme for the PF-ODE. This sampling step is possible provided that one can compute the actually intractable time-dependent score function $\score(\bx,t)=\bnabla_\bx\log\pdf_t(\bx)$ and sample from a known prior distribution $\pdf_T$. One approach is to use sliced score matching, which evaluates the projection of the score function onto a set of random directions \cite{song2020sliced}. However, denoising score matching techniques \cite{hyvarinen2005estimation,Song2019,song2020improved} have proven more effective by perturbing the initial samples in $\dataset$ with noise, thereby enhancing coverage of the data manifold and simultaneously defining a diffusion process governed by the chosen noise schedule. The score can be modeled using a time-dependent function $\score_\btheta(\bx_t, t)$, and the corresponding loss function to be minimized in explicit score matching is given by:
\begin{equation}
\Loss_{\text{SM}}(\btheta)=
    \demi \int_{0}^{T} 
    \esp_{\bx_{t}\sim \pdf_t(\bx_{t})}
    \left\{\llnorm{\score_{\btheta}(\bx_t,t) - \bnabla_\bx \log \pdf_t(\bx_t)
    }_{2}^2\right\} \id t\,,
\end{equation}
However, it is shown in \cite{vincent2011connection} that the following (weighted) loss function in denoising score matching can equivalently be minimized:
\begin{equation} \label{unconditional_obj}
\Loss_\text{DSM}(\btheta) =
    \demi\int _0 ^T \lambda(t) \, \esp_{\bx_0\sim \pdf_0(\bx_0)}\esp_{\bx_t\sim\pdf_{t|0}(\bx_t |\bx_0)}\left\{\norm{\score_\btheta(\bx_t,t)-\bnabla_\bx\log\pdf_{t|0}(\bx_t|\bx_0)}_{2}^{2}\right\} \id t \,,
\end{equation}
where $t\mapsto\lambda(t)$ is a weighting function (usually dependent on the noise variance) that can be set to the trace $\trace\,\bSigma$ of $\bSigma$ for example \cite{SMSDE} to obtain a likelihood weighting function \cite{huang2021variational,kingma2021variational,likelihoodFn} for an optimal training strategy, though other choices are possible \cite{song2020improved,karras2022elucidating,zheng2023improved} depending on the application. Notably, \eref{unconditional_obj} requires an explicit formulation of the transition kernel (transition probability) $\smash{\pdf_{t|0}(\bx_t|\bx_0)}$ associated with the forward SDE \eqref{SDE}. In practice, this generally implies selecting a linear drift $\bmu(\xv,t)=\mu(t)\xv$, and drift and diffusion coefficients that depend solely on time to ensure tractability. 
In the classical example of an Ornstein-Uhlenbeck process with $\mu(t)=-\demi$ and $\bsigma(t)=\Iv$, the score function reads:
\begin{equation*}
\score(\xv,t)=-\frac{\xv-\iexp^{-\frac{t}{2}}\esp\{\xv_0\lvert\xv_t=\xv\}}{1-\iexp^{-t}}\,,\quad t>0\,,
\end{equation*}
and $\denoiser(\xv,t)=-\sqrt{1-\iexp^{-t}}\score(\xv,t)$ is nothing but the continuous version of the denoiser in \eref{eq:denoising}, of which objective is actually a discrete version of \eref{unconditional_obj} with this reparameterization. Although multiple training strategies have been proposed to optimize $\smash{\Loss_\text{DSM}}$ \cite{song2020improved}, we adopt a random search approach due to the significant influence of numerous hyperparameters, particularly when incorporating regularization and sampling techniques.

\subsection{Autoregressive formulation and conditional score-matching objective}\label{sec:score-matching}

Among machine learning applications, solving PDEs is a particularly challenging yet constantly evolving task. Different strategies exist, including single-step predictions, next-step forecasting, and partial reconstructions. While many diffusion models are designed for direct or one-step simulations, achieving full PDE simulations remains the most demanding task. A few approaches have demonstrated impressive results \cite{lippe2023pde,du2024confild,ruhling2023dyffusion}, often paying the price of slow inference, due to the intrinsic complexity of the problem, limiting their applicability as plug-and-play surrogate models. A generic PDE system can be cast into the following equations:
\begin{equation}\label{eq:PDE}
\begin{cases}
\partial_{\tau}\bx+ \bs{\mathcal{F}} \left(\rv, \tau, \bx, \bnabla_\rv \bx, \dots \right)=
\mathbf{S}\left(\rv, \tau, \bx, \bnabla_\rv \bx, \dots \right)\,,
& \text{in } \Omega \times (0,\Tphysics)\,, \\ 
\BC\left(\rv, \tau, \bx, \bnabla_\rv \bx, \dots \right) = \zerov \,,
& \text{on } \partial\Omega \times (0,\Tphysics)\,, \\ 
\IC(\bx) = \bx(\rv, 0) \,,
& \text{in } \Omega \,,
\end{cases}
\end{equation}
where $\bx=(u, v, \Pres, \density \dots)\in\Rset^\ndim$ denotes a tensor-valued unknown frame representing a set of semi-discretized physical fields (\emph{e.g.}, in computational fluid dynamics, spanwise and streamwise velocities $u$ and $v$ in two dimensions, pressure $\Pres$, density $\density$...) at position $\rv \in \Omega \subset \Rset^d$ and time $\tau \geq 0$, $\bs{\mathcal{F}}$ is a nonlinear operator acting on $\bx$ and its spatial derivatives, $\bft{S}$ is a source term encapsulating external phenomena, and $\BC$ and $\IC$ define the boundary and initial conditions of the problem, respectively. To comply with diffusion model implementation, we adopt a double-index notation $\bx_t^\tau := \bx_t(\cdot,\tau)$ to emphasize the distinction between the physical timeline $\tau$ of \eref{eq:PDE} (see \ref{Phi time notation} for additional clarifications), and the diffusion process timeline $t$ of the generative diffusion models of \eref{eq:BSDE} or \eref{eq:PF-ODE}. In a probabilistic sense, the solution of \eqref{eq:PDE} at a given physical time $\tau$, provided with initial conditions $\IC$ and boundary conditions $\BC$, may be interpreted as a state $\hat{\bx}_{0}^{\tau}$ that maximizes the following conditional probability:
\begin{equation*}
    \hat{\bx}_{0}^{\tau}\in\arg\max_{\bx_{0}^{\tau}}\pdf_0 \left(\bx_{0}^{\tau}\,\middle|\,\tau,\IC(\bx_{0}^{0}),\BC(\bx_{0}^{\tau})\right)\,,
\end{equation*}
where $p_0$ stands for the probability density of the solution of \eqref{eq:PDE} conditioned on $\tau$, $\IC$, and $\BC$. This formulation highlights the fully predictive nature of the addressed problem, in contrast to the guiding constraints introduced by virtual observation inclusion as proposed by \citep{OBS_gen}. Consequently, maintaining simulation stability within this probabilistic framework proves particularly challenging. This is why many existing approaches tackle this issue by training a joint latent model, enriched with conditional information to strengthen the guidance impact on the prediction. Building on a provably efficient autoregressive framework, \citet{benchmarking} among others~(see for example \cite{shysheya2024conditional,GenCast25}) have proposed an approach that conditions the model on previous physical states. Indeed, the full discretization in time and space of \eref{eq:PDE} typically reads:
\begin{equation*}
\bx^\tau=\fv(\bx^{\tau-1},\bx^{\tau-2},\dots\bx^{\tau-m};\eta)\,,
\end{equation*}
where $\fv$ pertains to the discretization schemes used to solve \eqref{eq:PDE} and, say, generate the dataset $\dataset$, $\eta$ stands for all physical parameters, and $m$ is the number of past states considered in the time-marching integration scheme. In the proposed framework, this time-discretization scheme must be intended as a sampling process: the solution state $\bx^\tau$ generated at time $\tau$, starting from past states $\bx^{\tau-1},\bx^{\tau-2},\dots\bx^{\tau-m}$ (that are instances taken from the dataset or past predictions) is predicted as the one that maximizes a conditional likelihood. This predicted state is then used as a conditioning input to infer subsequent states, and the process continues iteratively. Efficient discretization schemes in DNS are typically fourth-order Runge-Kutta (RK) methods in time and fourth-order finite volumes in space \cite{DNS_base,SCH22}, but the intermediate time steps in RK methods are basically not provided in the datasets. Therefore the conditioning inputs may be limited to the most recent past solution in those datasets, setting $m=1$. This approach finally yields the maximization objective \cite{benchmarking,GenCast25,MUC25}:
\begin{equation}\label{eq:conditional_objective}
\hat{\bx}_{0}^{\tau}\in\arg\max_{\bx_{0}^{\tau}} ~ \pdf_{0} \left(\bx_{0}^{\tau}\,\middle|\,\bx_{0}^{\tau-1},\eta \right)\,, \quad \bx_{0}^{0} \in \dataset\,.
\end{equation}
For conciseness, we omit the dependency on the parameters $\eta$ in the remainder of the paper. As presented in \sref{continuous setting}, solving the backward SDE allows for the unconditional generation of new samples, $\hat{\bx}_0 \sim \pdf_\idata(\bx_0)$. To construct a conditional objective, several strategies have been proposed to incorporate conditioning into the sampling process of $\pdf_0(\bx_{0}^{\tau}\lvert\bx_{0}^{\tau-1})$, either by adapting the neural network architecture \cite{shysheya2024conditional} or by applying score-matching decomposition. 

While post-training conditioning can be addressed by exploiting the Bayesian properties of the probability density function \cite{hong2024score}, or by introducing a dedicated conditioner network \cite{jacobsen2025cocogen, gao2025generative}, a more recent alternative eliminates the need for an explicit conditioner through classifier-free guidance \cite{ho2022classifierfree,classifier_free}. In this approach, the model is jointly trained with and without conditioning information, allowing it to modulate the influence of the conditioning signal at inference time without a separate guidance model. Rather than adopting these strategies, we opt for the simpler design in which the model directly integrates both data and conditioning inputs \cite{benchmarking,dasgupta2024conditional} within a single training phase (see \sref{sec:UNet} for a detailed implementation). This streamlines the architecture and training process, although it does not guarantee superior performance. In the autoregressive setting considered, this choice favors fast training and efficient inference, deliberately avoiding explicit Bayesian formulations. 

One approach to obtain such conditional score model is to concatenate the encoded condition to the input, assuming they share the same shape. Thanks to \citet{batzolis2021conditional}, we can adapt the unconditional loss function for the conditional density $\pdf_{0}(\bx_{0}^{\tau}\lvert\bx_{0}^{\tau-1})$ and train a conditional score model $\score_\btheta(\bx_t^\tau,\bx_0^{\tau-1},t)$ with the loss:
\begin{equation}\label{eq:cSM}
 \Loss_\text{cSM}(\btheta) =
\demi\int_{0}^{T}
\lambda(t) \esp_{\bx_{t}^{\tau},\bx_{0}^{\tau-1} \sim p_{t}(\bx_{t}^{\tau}, \bx_{0}^{\tau-1})}
\left\{\llnorm{\score_\btheta(\bx_t^\tau,\bx_0^{\tau-1},t) - \bnabla_\bx \log p_{t\lvert 0}(\bx_{t}^{\tau}\lvert\bx_{0}^{\tau-1})}_{2}^2\right\} \id t\,,
\end{equation}
or equivalently:
\begin{equation}\label{eq:cDSM}
\Loss_\text{cDSM}(\btheta) =  
\demi\int_{0}^{T}
\lambda(t)\esp\repeatindex{
{\bx_{0}^{\tau},\bx_{0}^{\tau-1} \sim \pdf_{0}(\bx_{0}^{\tau}, \bx_{0}^{\tau-1})},
{\bx_{t}^{\tau}\sim \pdf_{t\lvert 0}(\bx_t^\tau\lvert\bx_0^\tau)}
}
\left\{\llnorm{\score_\btheta(\bx_t^\tau,\bx_0^{\tau-1},t) - \bnabla_{\bx} \log \pdf_{t \lvert 0}(\bx_{t}^{\tau} \lvert \bx_{0}^{\tau-1})}_{2}^2 \right\} \id t\,.
\end{equation}
Drawing samples $\bx_{0}^{\tau},\bx_{0}^{\tau-1} \sim \pdf_{0}(\bx_{0}^{\tau},\bx_{0}^{\tau-1})$ is simply done by jointly picking at random a physical state in the dataset and its past frame, assuming the dataset is large enough to consider the states $\bx$ as random samples from the solution space. To cover all paths, an average is made over all simulation time steps $\tau \in]0,\Tphysics]$ (ignoring the initial conditions at $\tau=0$) which gives the final Autoregressive-Modeling (AM) training score function:
\begin{equation}\label{eq:AM}
    \Loss_{\text{AM}}(\btheta) =
\demi \int_1^\Tphysics \int_{0}^{T}
\lambda(t)\esp\repeatindex{
{\bx_{0}^{\tau},\bx_{0}^{\tau-1} \sim \pdf_{0}(\bx_{0}^{\tau},\bx_{0}^{\tau-1})},
{\bx_{t}^{\tau}\sim \pdf_{t \lvert 0}(\bx_{t}^{\tau} \lvert \bx_{0}^{\tau})}
}
\left\{\llnorm{\score_\btheta(\bx_t^\tau,\bx_0^{\tau-1},t)
- \bnabla_{\bx} \log \pdf_{t \lvert 0}(\bx_{t}^{\tau} \lvert \bx_{0}^{\tau-1})}_{2}^2 \right\} \id t\id\tau\,.
\end{equation}
Again, in \eref{eq:cSM} through \eref{eq:AM} $t\mapsto\lambda(t)$ is a weighting function along the diffusion timeline $t$.

\section{Datasets and implementation}
\label{sec:data_and_archi}

\subsection{PDE systems}\label{sec:PDEsystems}

We first describe in detail the various PDE systems, all formulated under the general framework of \eqref{eq:PDE}, which are investigated using the auto-regressive, score-based diffusion model introduced in the previous section. Each system presents distinct physical challenges and flow characteristics, serving as diverse test cases for evaluating the robustness and generalization of the proposed approach. In \sref{sec:JHTBD}, we examine the case of a K\'arm\'an vortex street generated by the flow past a circular cylinder in two dimensions, a configuration well known for its periodic vortex shedding and complex wake dynamics. In \sref{sec:TRL}, we extend the analysis to a more complex scenario involving a turbulent radiative layer, still within a two-dimensional setting, characterized by strong multiscale interactions between turbulence and radiative transfer phenomena. Finally, in \sref{sec:MHD}, we consider a fully three-dimensional flow field coupled with magnetodynamic effects, representing a magnetohydrodynamic (MHD) system where the interplay between fluid motion and magnetic fields introduces additional layers of complexity to the predictive modeling task.

\subsubsection{Compressible transonic flow past a cylinder (transonic cylinder)}\label{sec:JHTBD}

Fluid mechanics problems arise across a wide range of disciplines. A classic example is the K\'arm\'an vortex street, which forms behind a cylinder placed in a fluid flow. This phenomenon has been extensively studied in the incompressible regime in particular. However, the compressible case deviates significantly from its incompressible counterpart and poses a considerable challenge, especially at high Reynolds numbers $\Reynolds$. While \citet{benchmarking} have tackled fluid prediction at high Reynolds numbers, the greater difficulty they encountered stems from the compressible regime, especially at transonic speeds with $\Reynolds = 10^4$. This regime is characterized by the emergence of shock waves, leading to high nonlinearities and a chaotic behavior. In this work, we focus on the more challenging task of extrapolation, using a simple, dedicated architecture. In contrast to the diverse network designs explored by \citet{benchmarking}, our approach focuses exclusively on optimizing the SDE formulation to yield a robust, lightweight, and plug-and-play solution. The dataset is composed of compressible transonic flow simulations\footnote{\href{https://github.com/tum-pbs/autoreg-pde-diffusion}{https://github.com/tum-pbs/autoreg-pde-diffusion}.} generated using the SU2 solver \cite{benchmarkingSolver}, with the training set $\dataset$ being made of simulation for Mach numbers $\Mach\in[0.53, 0.63] \cup [0.69, 0.90]$. The governing equations read:
\begin{equation}
\begin{cases}
\partial_\tau\density + \bnabla_\rv \cdot (\density\uv) = 0\,, \\
\partial_\tau(\density\uv) +\bnabla_\rv \cdot(\density\uv\otimes\uv-\stress) = \zerov \,, \\
\partial_\tau(\density E)+\bnabla_\rv \cdot((\density E - \stress)\uv+\heat) = 0 \,,
\end{cases}
\end{equation} 
where $\uv(\rv,\tau)$ is the fluid velocity field at position $\rv \in \Omega \subset \Rset^d$ and time $\tau\geq 0$, $\density(\rv,\tau)$ is the fluid density, $E(\rv,\tau)$ is the total energy per unit mass, and $\heat(\rv,\tau)$ is the heat flux vector due to thermal conductivity. Also $\otimes$ stands for the usual tensor product such that $(\bft{a}\otimes\bft{b})\bft{c}=(\bft{b}\cdot\bft{c})\bft{a}$ for any vectors $\bft{a},\bft{b},\bft{c}$ in $\Rset^d$ with the scalar product $(\cdot)$. For an isotropic Newtonian fluid, the stress tensor is $\stress = -\Pres\Iv + \lambda(\trace\,\bft{D})\Iv+2\mu\bft{D}$, where $\Pres(\rv,\tau)$ is the pressure field, $\lambda$ and $\mu$ are the bulk and dynamic viscosities of the fluid, and:
\begin{equation}\label{eq:strainvelocity}
\bft{D}=\demi(\bnabla\otimes\uv+(\bnabla\otimes\uv)^\itr)
\end{equation}
is the symmetrized velocity rate tensor of which trace is $\trace\,\bft{D}=\bnabla\cdot\uv$. The aim is to reconstruct flows in an extrapolation set of Mach numbers $\Mach\in[0.50, 0.52]$ for a total time lapse of $\Tphysics = 60$. The later results are displayed on \fref{fig:JHTDB_compa_fields} and \fref{fig:JHTDB_spectra} for three test cases corresponding to $\Mach=0.50$ (test case 0), $\Mach=0.51$ (test case 1), and $\Mach=0.52$ (test case 2) in two dimensions ($d=2$). Note that the images are downsampled to a $\ndim=128 \times 64$ format for the training and testing. This is done to save up time and energy considering the sheer number of cases to test.

\subsubsection{Turbulent radiative layer (TurbRad)}\label{sec:TRL}

In the pursuit of a unifying benchmark dataset for physics, "the Well" database \cite{theWell} encompasses a broad spectrum of scenarios and poses a compelling challenge in the field of fluid dynamics. In this context, the Turbulent Radiative Layer--2D (TurbRad) case\footnote{\href{https://polymathic-ai.org/the_well/datasets/turbulent_radiative_layer_2D/}{https://polymathic-ai.org/the$_-$well/datasets/turbulent$_-$radiative$_-$layer$_-$2D/}} \cite{2DradLay} introduces a mixing phenomenon between hot and cold gases, in addition to the inherent compressibility of the fluid, and offers an interesting scenario for the diffusion model. The governing equations read:
\begin{equation}
\begin{cases}
\partial_\tau\density + \bnabla_\rv \cdot (\density\uv) = 0 \,, \\
\partial_\tau(\density\uv) + \bnabla_\rv \cdot (\density\uv\otimes\uv+ \Pres\Iv) = \zerov \,, \\
\partial_\tau\Energy + \bnabla_\rv \cdot \left((\Energy + \Pres)\uv \right) = -\frac{\Energy}{\tau_\text{cool}} \,,
\end{cases}
\end{equation}
where $\uv(\rv,\tau)$ is the velocity field at position $\rv \in \Omega \subset \Rset^2$ and time $\tau\geq 0$, $\density(\rv,\tau)$ is the density, $\Pres(\rv,\tau)$ the pressure field, $\Energy(\rv,\tau)=\frac{\Pres(\rv,\tau)}{\gamma-1}$ is the total energy per unit volume with $\gamma=\frac{5}{3}$, and $\smash{\tau_\text{cool}}$ is the cooling time. In a similar vain, the aim is to solve the flow in a $\ndim=192 \times 64$ spatial resolution space and a time lapse of $\Tphysics = 60$ (similarly to \sref{sec:JHTBD}, the format has been downsampled to illustrate the applicability of the model and not engage too much computational costs). The results are displayed on \fref{fig:turbRadPred} and \fref{fig:turbRadSpectra} for the three test cases corresponding to $\smash{\tau_\text{cool}}=0.03$ (test case 0), $\smash{\tau_\text{cool}}=0.06$ (test case 1), and $\smash{\tau_\text{cool}}=0.1$ (test case 2). This dataset presents a considerable challenge due to abrupt variations in the fields, discontinuities in the velocity components (in contrast to the smooth vector fields observed in lower-speed flows), and large discrepancies in the data distribution (both inter-field and intra-field).

\subsubsection{Magnetohydrodynamics 64 (MHD)}\label{sec:MHD}

This selected dataset aims to illustrate the 3D extrapolation capabilities of the diffusion model and demonstrate that, despite its conceptual simplicity, the 2D network can be extended to a 3D flow with minimal effort. In this context, the MHD$_-$64 (or magnetohydrodynamics 64) case\footnote{\href{https://polymathic-ai.org/the_well/datasets/MHD_64/}{https://polymathic-ai.org/the$_-$well/datasets/MHD$_-$64/}} \cite{MHD64} provided by "the Well" database \cite{theWell} fulfills this role by providing a 3D flow coupled with magnetodynamics. This dataset is governed by the ideal compressible MHD equations, which read:
\begin{equation}
\begin{cases} 
\partial_\tau\density + \bnabla_\rv\cdot(\density\uv)= 0\,, \\
\partial_\tau(\density\uv) + \bnabla_\rv\cdot(\density\uv\otimes\uv +\Pres\Iv) + \frac{1}{\mu_0}\bft{B}\times(\bnabla_\rv\times\bft{B}) = \zerov \,, \\
\partial_\tau\bft{B} - \bnabla_\rv \times (\uv \times \bft{B}) = \zerov \,,
\end{cases}
\end{equation}
where $\uv(\rv,\tau)$ is the velocity field at position $\rv \in \Omega \subset \Rset^3$ and time $\tau\geq 0$, $\density(\rv,\tau)$ is the fluid density, $\Pres(\rv,\tau)$ is the gas pressure, $\bft{B}(\rv,\tau)$ is the magnetic field, and $\mu_0$ is the magnetic permeability. The raw format of the fields being $64^3$, we also downsample them to $\ndim=32^3$ to illustrate the capabilities of the model. The results are displayed on \fref{fig:MHD_pred} and \fref{fig:MHD_spectrum} for the three test cases corresponding to sonic Mach number $\Mach_s=0.5$ (test case 0),  $\Mach_s=0.7$ (test case 1), and $\Mach_s=1.5$ (test case 2) under a fixed Alfv\'en (magnetic) Mach number of $\Mach_A = 0.7$. These settings span subsonic to mildly supersonic regimes within a sub-Alfv\'enic flow, providing a controlled framework to evaluate the model's ability to capture turbulence across varying compressibility levels while preserving a strong magnetic influence. 

Due to the substantial computational cost of generating full temporal predictions, the evaluation is limited to $\mathcal{T}=20$ predicted time steps per case. In addition to the emergence of realistic 3D turbulence, the sheer size of the dataset and the coupling with additional dynamics make it a genuine challenge for diffusion models, further highlighting their versatility. This dataset presents an additional difficulty, not only due to the large prediction space, which might significantly amplify the propagation and accumulation of errors, but also because it includes additional fields with different dynamics, requiring distinct statistical measures.

\subsection{Implementation}

We now describe the implementation of the proposed auto-regressive, score-based diffusion model. In \sref{sec:UNet} we first briefly outline the network architecture of the denoiser. Then in \sref{sec:SDEs} we summarize the selected SDE models (drift and diffusion coefficients) whose performances will be analyzed subsequently in \sref{sec:Results}. In \sref{sec:regularization} the regularization procedure that has been developed to train the denoiser is detailed.

\subsubsection{Network architecture}\label{sec:UNet}

A U-Net denoiser architecture \cite{UNet}, enhanced with attention mechanisms inspired by the Transformer framework \cite{vaswani2023}, is employed to address the reconstruction task. This choice balances computational efficiency and denoising performance, leveraging architectural components that have been extensively optimized in the literature. The model remains lightweight, easy to implement, and well-adapted to high-dimensional generative settings. To extend this architecture to 3D volumetric data, we adopt the implementation provided by the Hugging Face Diffusers library\footnote{\href{https://huggingface.co/docs/diffusers/api/models/unet3d-cond}{https://huggingface.co/docs/diffusers/api/models/unet3d-cond}}, which offers a robust baseline for 3D U-Net models. This adaptation enables effective handling of 3D flow fields while maintaining the original design's efficiency.

\begin{figure}[t]
\centering
\includegraphics[width=\textwidth]{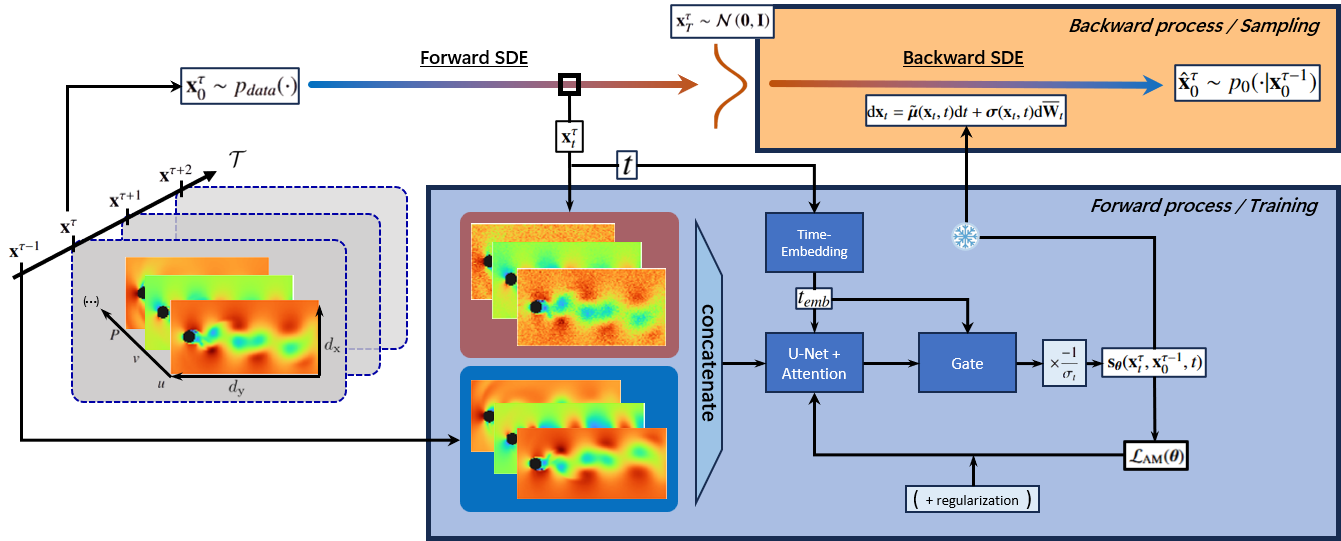}
\vspace{-1em}
\caption{Schematic representation of the model architecture, training and sampling procedure.}\label{fig:architecture}
\end{figure}

Following the conditioning and autoregressive framework proposed in \citet{shysheya2024conditional} and \citet{benchmarking}, the architecture (see \fref{fig:architecture}) is modified to support multichannel inputs, allowing the simultaneous processing of multiple physical variables and the conditioning information. The conditioning is implemented via channel-wise concatenation, a strategy introduced in \citet{isola2017image}. Prior to concatenation, the conditioning inputs are encoded using a series of U-Net ResBlocks \cite{DDPM,SMSDE}. This approach effectively integrates auxiliary information while preserving the spatial coherence and hierarchical structure required for generation. However, high noise variance during training is known to induce instabilities and hinder performance \cite{karras2022elucidating}. To mitigate this issue without altering the core network structure, we train the model to predict the clean target, and introduce a scalable correction mechanism at inference time. This post-processing step compensates for residual noise-related discrepancies and improves robustness, particularly in high-noise regimes:
\begin{equation}
\denoiser_\btheta(\xv_i,t) \equiv \alpha_\btheta(t_\text{emb}) \cdot \texttt{U-Net}(\xv_i, t)
\end{equation}
where skip connections are implicitly included in the architecture, $\alpha_{\btheta}(t) = \texttt{sigmoid}(Wt + b)$ is a learnable scale factor with weight $W$ and bias $b$, and $t_\text{emb}$ is a vector that encodes the current diffusion step or noise level \cite{DDPM,vaswani2023}. This allows the network to modulate the magnitude of its correction based on the noise level, enabling it to suppress unnecessary updates when the noise is low. The same model is used for the score function $\score_\btheta$ in \eqref{eq:AM}.

\subsubsection{SDE models}\label{sec:SDEs}

\begin{table}[t]
        
        


    \centering
    \renewcommand{\arraystretch}{2.} 

    \begin{tabular}{cccc}
        \hline
        \textbf{SDE type} & \textbf{Formulation} & {$\bmu_t(\xv_0)$} & {$\bSigma_t(\xv_0)$} \\
        \hline
        
        VP & $ \id\bx_{t}=-\demi\beta(t)\bx_t \id t + \sqrt{\beta(t)}\,\id\bW_{t}$ &
        $\bx_{0}\iexp^{-\demi\int_0^t \beta(s)\id s}$ & $\parenthesis{1 - \iexp^{-\int_0^t\beta(s)\id s}} \Iv$ \\
        \hline
        
        sub-VP & $ \id\bx_t=-\demi\beta(t)\bx_t \id t + \sqrt{\beta(t)\left(1-\iexp^{-2\int_0^t\beta(s)\id s}\right)}\,\id\bW_t$ 
        & $\bx_0 \iexp^{-\demi\int_0^t \beta(s)\id s}$ & $\parenthesis{1 - \iexp^{-\int_0^t \beta(s) \id s}}^{2} \Iv$ \\
        \hline

        VE & $ \id\bx_t = \sqrt{\frac{\id[\sigma^2(t)]}{\id t}} \, \id\bW_t$ &
        $\bx_0$ & $\brackets{\sigma^2(t) - \sigma^2(0)} \Iv$ \\
        \hline

    \end{tabular}

    \caption{SDE formulations proposed by \citet{SMSDE} and associated transition kernel parameters. The transition kernel is $\pdf_{t\lvert 0}(\bx_t\lvert\bx_0)=\Normal(\bx_t;\bmu_t(\xv_0),\bSigma_t(\xv_0))$ where $\bmu_t(\xv_0)=\esp\{\bx_t\lvert\bx_0\}$ and $\bSigma_t(\xv_0)=\Var\{\bx_t\lvert\bx_0\}$, for $\Var$ being the variance.}
    \label{tab:SDEs}
\end{table}

The diffusion model formulated in \eref{SDE} offers substantial flexibility through the parametrization of its drift and diffusion coefficients. This versatility allows tailoring the SDE framework to replicate and generalize several prominent generative modeling approaches \cite{SMSDE} (see \tref{tab:SDEs} for a synthesis of these formulations). In particular, three distinct SDE parameterizations have emerged as especially relevant due to their structural and theoretical correspondence with state-of-the-art generative models. By judiciously specifying the drift coefficient $\bmu$ and the diffusion coefficient $\bsigma$, the SDE can be configured to emulate well-established stochastic processes.

A classical example is the Ornstein-Uhlenbeck process, where the drift exhibits linear mean-reverting behavior and the diffusion is constant. Within the generative modeling context, an analogous formulation leads to the continuous-time limit of popular score-based and diffusion models. Specifically, this framework encompasses the variance-exploding SDE (VE SDE), introduced in the context of annealed Langevin dynamics \cite{Song2019}, where $\bsigma$ increases monotonically over time to facilitate large-scale exploration of the sample space. Conversely, variance-preserving SDEs (VP SDE), as employed in DDPMs \cite{DDPM}, maintain a controlled noise level with $\bsigma$ designed to preserve the overall variance of the data distribution throughout the forward diffusion process.

The third formulation, the sub-VP SDE, introduced in \cite{SMSDE}, derives from a further modification of the VP SDE paradigm. Its naming stems from the specific dependency of the diffusion coefficient $\bsigma$ on the temporal variance evolution, directly linked to the ordinary differential equation associated with the SDE dynamics \cite{sarkka2019applied}. This modification yields enhanced numerical stability and improved control over the signal-to-noise ratio at intermediate time steps.

\subsubsection{Regularization}\label{sec:regularization}

To further improve the temporal coherence of the predictions, an additional regularization strategy is proposed. Each sample being conditioned by its preceding states, we introduce a supplementary loss term that penalizes deviations related to the system’s energy, with the objective of enhancing time consistency. We begin by decomposing the flow velocity field $\uv$ in the PDE systems described in \sref{sec:PDEsystems} into a mean component $\Uv$ and a stochastic fluctuating component $\uv'$, such that $\uv'(\bft{r},\tau) = \uv(\bft{r},\tau) - \Uv(\bft{r},\tau)$, where $\esp\{\uv'(\bft{r},\tau)\} = \zerov$ by construction \cite{POP00}. The mean component is obtained by averaging $\bft{u}(\bft{r}, \tau)$ and $\bft{u}(\bft{r}, \tau - 1)$. This is justified by the rapid evolution of the flow, which implies a short Lyapunov time \cite{lya} (see \ref{sec:ftle_lyapunov_time} for application on the datasets). Since the system decorrelates quickly, randomness emerges even over such a short time interval, making this local averaging appropriate. The matrix-valued cross-correlation function of the velocity field fluctuations is then defined as:
\begin{equation*}
\Corr(\rv,\rv',\tau,\tau')=\esp\{\uv'(\rv,\tau)\otimes\uv'(\rv',\tau')\}\,.
\end{equation*}
It can be computed for the set of positions $\{\rv\}_{\rv\in \Omega}$ (excluding the obstacle if any, as in \emph{e.g.}, the case of the cylinder in \sref{sec:JHTBD}) and their associated velocity fields $\uv(\rv,\tau)$ from the training dataset $\dataset$. We take an equivalent form of a Fischer-like divergence on the cross-correlations of the velocity field as the regularization cost function:
\begin{equation*}  
\bb{D}\parenthesis{\uv,\hat{\uv}}=\esp_{\uv\in\dataset}\esp_{\rv,\rv'\in \Omega}\left\{\norm{\uv'(\rv,\tau)\otimes\uv'(\rv',\tau-1)-\hat{\uv}'(\rv,\tau)\otimes\hat{\uv}'(\rv',\tau-1)}_F^2\right\}\,,
\end{equation*}
where $\norm{\cdot}_F$ is the Frobenius norm, and $\hat{\uv}$ is the estimated velocity field retrieved during training. Indeed, since $\smash{\score_\btheta(\bx_t^\tau,\bx_0^{\tau-1},t)}\simeq - \smash{(\hat{\bx}_t^\tau - \bx_0^\tau)/\sigma_t^2}$ whenever $\bSigma_t=\sigma_t^2\Iv$ \cite{vincent2011connection}, we can retrieve $\hat{\uv}(\cdot,\tau)$ from the score. During the training phase, we consider $\hat{\uv}(\rv,\tau-1)\equiv\uv(\rv,\tau-1)$, imposing time consistency with the dataset rather than relying on a potentially inaccurate model prediction. This deliberate choice ensures that the temporal evolution of the velocity field remains physically grounded in the reference data and that the model focuses on learning proper conditional dynamics of the flow at each time step. We then rewrite:
\begin{equation*}
\begin{aligned}
\bb{D}\parenthesis{\uv,\hat{\uv}}
& =
\esp_{\uv\in\dataset}\esp_{\rv,\rv'\in\Omega}
\left\{\norm{\uv'(\rv,\tau)\otimes\uv'(\rv',\tau-1) - 
\hat{\uv}'(\rv,\tau)\otimes\uv'(\rv',\tau-1)}_F^2 \right\} \\
& =
\esp_{\uv\in\dataset}\esp_{\rv,\rv'\in\Omega}
\left\{\norm{\brackets{\uv'(\rv,\tau) - \hat{\uv}'(\rv,\tau)}
\otimes\uv'(\rv',\tau-1)}_F^2 \right\} \,.
\end{aligned}
\end{equation*}
However $\norm{\bft{a}\otimes\bft{b}}_F^2 = \sum_{i,j} (a_i b_j)^{2} = \norm{\bft{a}}_2^2\norm{\bft{b}}_2^2$, so we have:
\begin{equation*}
\begin{aligned}
\bb{D}\parenthesis{\uv,\hat{\uv}} 
& \simeq 
\frac{1}{\abs{\Omega}^2}\esp_{\uv\in\dataset}\left\{\sum\limits_{\rv,\rv'\in \Omega}
\norm{\uv'(\rv,\tau) - \hat{\uv}'(\rv,\tau)}_2^2\norm{\uv'(\rv',\tau-1)}_2^2\right\} \\
& = 
\frac{1}{\abs{\Omega}^2}\esp_{\uv\in\dataset}\left\{
\sum\limits_{\rv\in \Omega}\norm{\uv'(\rv,\tau) - \hat{\uv}'(\rv,\tau)}_2^2 \sum\limits_{\rv'\in\Omega}\norm{\uv'(\rv',\tau-1)}_2^2\right\}\,.
\end{aligned}
\end{equation*}
It is then proposed to add this very cost function to the non-regularized loss function $\Loss_\text{AM}$ of \eref{eq:AM} to further enhance stability at the very low cost of a global sum over space and time:
\begin{equation}
\Loss_\text{rAM}(\btheta) = \Loss_\text{AM}(\btheta) +
\lambda_w \int_1^\Tphysics \bb{D}\parenthesis{\uv,\hat{\uv}} \id\tau\,.
\end{equation}
This augmented loss (with some weighting constant $\lambda_w>0$ adapted to each SDE scheme in \tref{tab:SDEs} and PDE system) will be compared to the vanilla loss function $\Loss_{\text{AM}}$.

\section{Results}\label{sec:Results}

When comparing a model’s prediction to a reference simulation in fluid dynamics, it is crucial to first clarify the objective and identify the key statistical features of the flow that the diffusion model should capture. While the Navier-Stokes equations are deterministic and theoretically admit finite-energy (weak) solutions for all time given well-defined initial and boundary conditions \cite{ROB20}, their highly chaotic behavior in turbulent regimes makes them extremely sensitive to small perturbations. As a result, even slight differences in input or numerical approximation can lead to significantly different flow realizations. In such a context, evaluating a model's performance requires focusing on statistical similarity rather than exact pointwise agreement. 

\subsection{Training and common evaluation metrics}

\begin{table}[t]
\centering
{\small
\begin{tabular}{cc cccc cccc cccc}
\hline
{\textbf{SDE type}} & {\textbf{Regularization}} & \multicolumn{4}{c}{Transonic cylinder} & \multicolumn{4}{c}{TurbRad} & \multicolumn{4}{c}{MHD} \\ \cline{3-14} 
 &  & $\sigma_\imin$ & $\sigma_\imax$ & $\beta_{\text{min}}$ & $\beta_{\text{max}}$ & $\sigma_{\text{min}}$ & $\sigma_{\text{max}}$ & $\beta_{\text{min}}$ & $\beta_{\text{max}}$ & $\sigma_{\text{min}}$ & $\sigma_{\text{max}}$ & $\beta_{\text{min}}$ & $\beta_{\text{max}}$ \\ \hline
VP & False     & - & -  & 0.01 & 5    & - & - & 0.1 & 25    & - & - & 0.1  & 30 \\
sub-VP & False & - & -  & 0.35 & 30   & - & - & 0.4  & 30    & - & - & 0.25 & 30 \\
VE & False     & 0.04 & 8 & - & -     & 0.15 & 4 & - & -  & 0.1 & 6 & -  & -  \\
\hline
VP & True      & - & - & 0.39 & 5.6  & - & - & 0.1 & 27     & - & - & 0.15 & 7 \\
sub-VP & True  & - & - & 0.4 & 28    & - & - & 0.25 & 30     & - & - & 0.1 & 12  \\
VE & True      & 0.05 & 8 & - & -    & 0.05 & 5 & - & -     & 0.01 & 4 & - & -  \\
\hline
\end{tabular}}
\caption{Optimal training parameters for the different types of SDE without and with regularization.}
\label{tab:sde_parameters}
\end{table}

While smaller models often benefit from faster convergence and reduced training complexity, the large number of hyperparameters involved introduces considerable computational challenges due to intricate interdependencies. Since no single configuration consistently yields optimal performance across all scenarios, a random search strategy is adopted to efficiently probe the high-dimensional hyperparameter space. The sampling ranges are guided by established heuristics and insights from prior work (see \sref{sec:data_and_archi}). Each candidate configuration is trained under consistent conditions, with a sufficient number of iterations to allow for convergence. To ensure fairness and efficiency, early stopping is systematically employed to avoid overfitting and limit unnecessary computations, along with a fixed upper bound on the number of training epochs to ensure reproducibility.

To assess and compare the performance across the various flow configurations and SDE formulations, a combination of evaluation metrics is used. These include both global error measures and statistical indicators that collectively capture reconstruction accuracy, structural coherence, and distributional similarity. Visualizations of predicted flow fields remain the primary diagnostic tool, given their relevance for identifying qualitative discrepancies in fluid dynamics. Representative snapshots are shown in the main figures, while other time series are provided in \ref{app:add_results} to support more detailed assessment and interpretation over a more complete time horizon. The optimal hyperparameter sets resulting from this process are presented in \tref{tab:sde_parameters}, offering a reproducible baseline for future studies. As a side note, each model contained approximately 16.6 million parameters ($\sim\,$66MB, with 99$\%$ being trainable). For the transonic cylinder and TurbRad cases, training was conducted on two NVIDIA A100 GPUs (each with 40GB HBM2e memory), using 4 CPU cores (Intel Xeon Gold 6230 20C, 2.1GHz) per thread and 60GB of RAM. Training time was about 1 hour per model for the transonic cylinder and approximately 2 hours for TurbRad. For the MHD model, training was performed on four NVIDIA Tesla P100 GPUs (each with 16GB memory), using 6 CPU cores per thread and 100GB of RAM, with each model requiring roughly 4 hours to train. Sampling was performed on a machine equipped with an NVIDIA GeForce RTX 3090 GPU (24GB VRAM). On this setup, generating a full time series for each case required approximately 10 minutes for Transonic cylinder, 30 minutes for TurbRad, and 1 hour for the MHD dataset.

\subsubsection{Averaged measures}

To assess the quality of the predicted flow fields, we employ a set of metrics that capture different aspects of model performance. These metrics are computed at each time step and then averaged over the entire temporal sequence (with an overbar $\taverage{(\cdot)}$ denoting the time average) to provide a global evaluation. When appropriate, the averaging is also applied across all predicted fields to enhance interpretability and reduce local variability, which is often observed with such flows. This yields a robust quantitative assessment framework that complements qualitative visual inspection, especially at late time steps where structural errors tend to be more pronounced.

We report the Mean Squared Error ($\MSE$), a widely used metric in image processing that quantifies the average squared difference between predicted and ground-truth flow fields. While coarse, it serves as a baseline indicator of reconstruction accuracy. To assess structural alignment and temporal coherence, we also compute the Pearson correlation coefficient ($\pearson$) \cite{BEN09}, which measures the linear correlation between predicted and reference values. This metric is particularly informative for evaluating the preservation of spatiotemporal patterns in the predicted dynamics. Additionally, we evaluate the Kullback–Leibler divergence ($\DKL$) between histograms of predicted and ground-truth field values. This information measure captures differences in the underlying statistical distributions, providing insight into how well the model reproduces the variability and structural characteristics of the flow. Taken together, these three metrics offer a comprehensive quantitative evaluation that complements visual diagnostics and supports a global interpretation of model performance.

\subsubsection{Energy spectrum}

For all three cases, a particularly insightful metric is the energy spectrum, which characterizes the distribution of energy across scales and serves as an indicator of the model’s capability to reconstruct the global dynamics. We evaluate the energy density spectrum $\kj\mapsto\FFT{E}(\kj,\tau)$ defined as the spherical average of the spatial Fourier transform $\FFT{\Corr}(\kv,\tau)$ (\emph{i.e.} the power spectral density) of the auto-correlation function of the velocity field fluctuations, namely:
\begin{equation}
\FFT{E}(\kj,\tau)=\demi\int_{\norm{\kv}=\kj}\trace\,\FFT{\Corr}(\kv,\tau)\id S=\frac{\kj^2}{2}\int_{S^{d-1}}\trace\,\FFT{\Corr}(\kj\unit{\kv},\tau)\id\mathit{\Omega}(\unit{\kv})\,,
\end{equation}
where $S^{d-1}$ stands for the unit sphere of $\Rset^d$ with the uniform probability measure $\mathit{\Omega}$, $\unit{\kv}=\frac{\kv}{\norm{\kv}}$ whenever $\norm{\kv}>0$, and:
\begin{equation*}
\FFT{\Corr}(\kv,\tau)=\frac{1}{(2\pi)^d}\int_{\Rset^3}\iexp^{-\ci\kv\cdot\rv'}\Corr(\rv,\rv+\rv',\tau,\tau)\id\rv'\,,
\end{equation*}
independently of $\rv$ in the homogeneous case. Note that the power spectral density may also be computed from the identity:
\begin{equation*}
\esp\{\FFT{\uv}'(\kv,\tau)\otimes\cjg{\FFT{\uv}'(\kv',\tau)}\}=(2\pi)^d\delta(\kv-\kv')\FFT{\Corr}(\kv,\tau)\,,
\end{equation*}
where $\cjg{\bft{a}}$ stands for the complex conjugate of $\bft{a}\in\Cset^d$, and:
\begin{equation*}
\FFT{\uv}'(\kv,\tau)=\frac{1}{(2\pi)^d}\int_{\Rset^3}\iexp^{-\ci\kv\cdot\rv}\uv'(\rv,\tau)\id\rv\,.
\end{equation*}
In the inertial subrange one expects $\smash{\FFT{E}(\kj,\tau)}\propto\smash{\kj^{-\frac{5}{3}}}$ independently of $\tau$, either in 2D or in 3D \cite{POP00}. Since the model outputs a full time series of velocity fields, computing spectra at each frame is redundant. We instead average across time ($\taverage{E}(\kj)=\frac{1}{\Tphysics}\sum_{\tau=1}^\Tphysics{\FFT{E}(\kj,\tau)}$) to obtain a representative spectral distribution that captures overall behavior while preserving clarity. For improved visual comparison, the spectra are plotted using different horizontal scales, with cutoffs applied to emphasize the relevant $\kj$ range and accommodate the varying magnitudes across cases.

In the context of chaotic flows, the energy spectrum typically spans several orders of magnitude across discrete wavenumbers $\{\kj_k\}_{k=1}^K$, especially in regimes involving inertial cascades. This wide dynamic range implies that absolute differences at high-energy (large-scale) modes can dominate traditional error metrics such as the $\MSE$, thereby masking discrepancies in lower-energy (small-scale) regions that may still carry essential physical information. To address this imbalance and provide a more scale-aware comparison, we adopt the $\log-\MSE$ defined as:
\begin{equation*}
\log-\MSE = \frac{1}{K} \sum_{k=1}^K \left[ \log\left( \taverage{E}^\text{GT}(\kj_k) + \varepsilon \right) - \log\left( \taverage{E}^\text{pred}(\kj_k) + \varepsilon \right) \right]^2,
\end{equation*}
where $\smash{\taverage{E}^\text{GT}}$ is the ground truth spectrum, $\smash{\taverage{E}^\text{pred}}$ is the predicted one and $\varepsilon$ is a small regularization constant to avoid numerical instability. This formulation emphasizes relative differences across each spectrum, thereby capturing both large and small scale discrepancies. In turbulence modeling, where spectral fidelity is key, the $\log-\MSE$ provides a more robust and interpretable measure of model performance than the raw $\text{MSE}$.

\subsection{Transonic cylinder} \label{sec:res_JHTDB}

\begin{figure}[t]
\centering
\includegraphics[width=\linewidth,keepaspectratio=True]{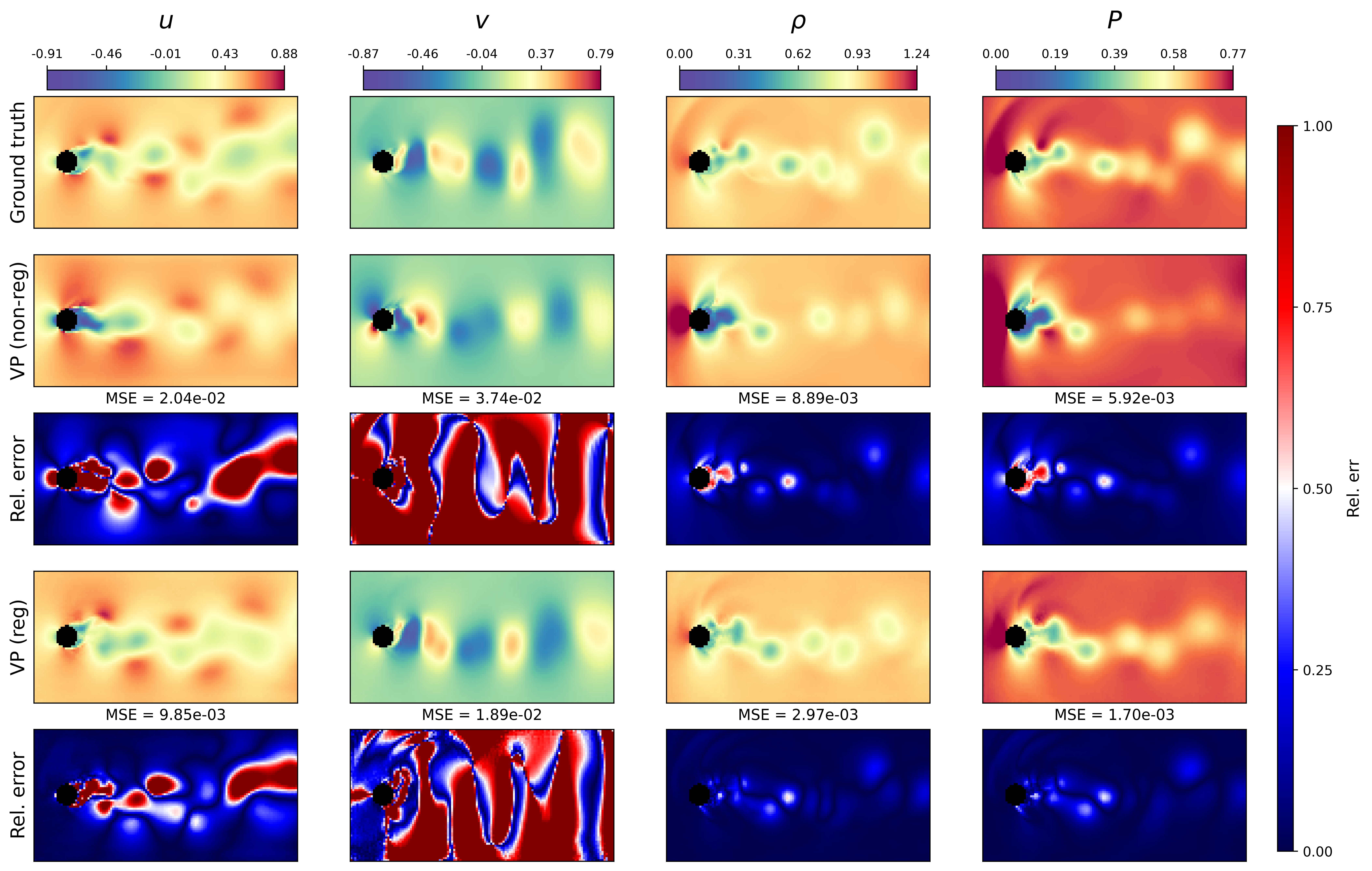}
\caption{Example test sample comparison at time $\tau = 60$ across fields between a VP-trained model without (non-reg) and with regularization (reg) for the transonic cylinder at $\Mach=0.52$ (test case 2). The mean squared error (MSE) and relative error (Rel. error) are also shown.}
\label{fig:JHTDB_compa_fields}
\end{figure}

\begin{figure}[t]
\centering
    \includegraphics[width=\linewidth]{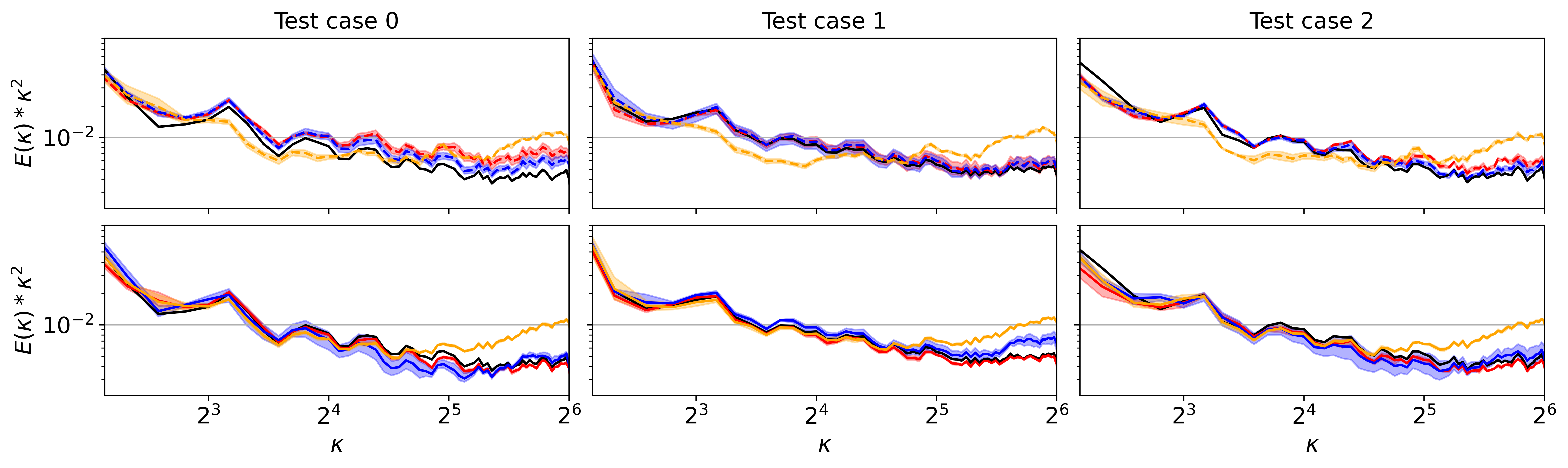}
    \vspace{-1em}
\caption{Time averaged spatial turbulent kinetic energy (spectra) of non-regularized (top, dashed lines) and regularized (bottom, solid lines) trained models for the transonic cylinder dataset. Shaded area shows the 5th to 95th percentile across sampled trajectories. \textcolor{black}{\rule[1.5pt]{0.5cm}{1pt}} ground-truth; \textcolor{red}{\rule[1.5pt]{0.5cm}{1pt}} VP SDE; \textcolor{blue}{\rule[1.5pt]{0.5cm}{1pt}} sub-VP SDE; \textcolor{orange}{\rule[1.5pt]{0.5cm}{1pt}} VE SDE.}
\label{fig:JHTDB_spectra}
\end{figure}

The cylinder case is characterized by relatively simpler dynamics compared to the TurbRad and MHD cases. the training process quickly converges after 400 epochs so the limit is set at 600 epochs. The relatively small size of the dataset (with $\ndim = 128 \times 64$) makes it for quick training and fast optimization.
The reconstruction of the energy spectrum in \fref{fig:JHTDB_spectra} highlights significant performance disparities across the SDE formulations. Among these, the regularized VP SDE consistently stands out, achieving high fidelity across the frequency range, while the VE formulation, regardless of regularization, struggles particularly at high frequencies. This shortcoming reflects an intrinsic limitation of VE-based dynamics in resolving fine-scale structures, an issue plausibly caused by the use of an exponential noise schedule. As discussed in \cite{SMSDE,kingma2021variational}, such schedules are particularly sensitive to hyperparameter selection and often demand careful tuning and extended training to achieve reliable performance, especially in regimes dominated by sharp gradients. Nevertheless, the introduction of regularization yields substantial improvements in the lower frequency range for all models, including VE, indicating that regularization plays a crucial role in enhancing the coherence of large-scale structures and stabilizing the learning dynamics.

These benefits are especially evident for the VP and sub-VP SDEs, whose spectral reconstructions are significantly improved under regularization. As shown quantitatively in \tref{tab:CylinderMetrics} and visually in \fref{fig:Cylinder_time_series} in \ref{app:time_series}, the regularized VP SDE outperforms all other variants across all evaluation criteria, including the lowest $\log-\MSE$, highest $\pearson$, and minimal $\MSE$. These metrics, which provide complementary insight into both the global and local prediction accuracy, support the claim that the VP SDE formulation offers the most robust and accurate temporal modeling capabilities as suggested by \cite{benchmarking}. This is further supported by qualitative comparisons in \fref{fig:JHTDB_compa_fields}, which depicts final-time predictions at $\tau = \Tphysics = 60$ for the transonic test case 2 at $\Mach = 0.52$. In this context, the regularized VP SDE clearly demonstrates its superiority when regularized, particularly in regions near solid boundaries and obstacles, where the flow is dominated by steep gradients and shock structures. The clarity of upstream shocks and preservation of coherent vortex patterns are especially noticeable, validating the advantage of regularization for fine-scale fidelity and structural preservation.

When turning exclusively to the sub-VP SDE, the results are more nuanced. While no uniform improvement is observed across all test cases, specific instances, such as test case 0, do exhibit a notable structural enhancement with regularization, as evidenced in \fref{fig:Cylinder_time_series}. A similar effect is observed for the VE SDE, where the unregularized model produces faded and incoherent fields, whereas the regularized counterpart recovers much of the lost structure, particularly in early and intermediate stages. The phenomenon of temporal degradation is more directly addressed in \fref{fig:Cylinder_time_series}, which displays a progressive loss of coherence in the predicted pressure field $\Pres$ over time. This deterioration is a challenge in autoregressive or iterative sequence prediction \cite{shysheya2024conditional} and demonstrates the inherent difficulty in preserving physical consistency over long horizons (even in relatively well performing models). Remarkably, the regularized VP SDE resists this degradation, maintaining coherent pressure and velocity structures throughout the timeline. This robustness confirms the stabilizing role of regularization not only in the spectral domain but also in the physical evolution of the fields. These findings resonate with the results of \cite{benchmarking}, which emphasized that low $\MSE$ across the timeline serves as a reliable demonstration of the extrapolative robustness of generative models, particularly in compressible transonic flow regimes where fine resolution is critical.

Finally, \fref{fig:U_mag} in \ref{app:general_compa} provides further evidence of this robustness by displaying the time evolution of the velocity magnitude $\norm{\uv}$. While most models eventually diverge or exhibit visible drift, the regularized versions consistently maintains coherent and physically plausible dynamics, especially at late times. This resiliency underscores its suitability for high-fidelity flow modeling tasks, confirming that among the tested formulations, the regularized VP SDE represents the most balanced and reliable approach, capable of simultaneously capturing multi-scale structure, maintaining long-term stability, and delivering accurate predictions across both spatial and temporal dimensions.

\begin{table}[t]
\centering
{
\begin{tabular}{cc ccc ccc ccc ccc}
\hline
\textbf{SDE} & \textbf{Reg.} & \multicolumn{3}{c}{$\taverage{\MSE} \downarrow$ ($10^{-3}$)} & \multicolumn{3}{c}{$\taverage{\pearson} \uparrow$ ($10^{-1}$)} & \multicolumn{3}{c}{$\taverage{\DKL} \downarrow$ ($10^{0}$)} & \multicolumn{3}{c}{$\log-\MSE \downarrow$ ($10^{-2}$)} \\ \cline{3-14} 
& & tc 0 & tc 1 & tc 2 & tc 0 & tc 1 & tc 2 & tc 0 & tc 1 & tc 2 & tc 0 & tc 1 & tc 2 \\ \hline
VP & False & 2.52 & 1.49 & 2.82 & 9.90 & 9.93 & 9.88 & 5.01 & 1.98 & 5.25 & 2.87 & 0.27 & 0.82 \\
sub‑VP & False & 1.77 & 1.26 & 4.94 & 9.92 & 9.94 & 9.78 & 5.02 & \textbf{1.88} & 5.34 & 0.94 & 0.29 & \textbf{0.27} \\
VE & False & 7.45 & 8.67 & 8.17 & 9.67 & 9.61 & 9.64 & 5.23 & 2.50 & 5.49 & 5.84 & 5.16 & 4.77 \\
\hline
VP & True & \textbf{1.67} & \textbf{1.06} & 3.10 & \textbf{9.93} & \textbf{9.95} & 9.86 & 5.07 & 2.12 & \textbf{5.13} & \textbf{0.20} & \textbf{0.09} & 0.30 \\
sub‑VP & True & 2.40 & 2.41 & 3.55 & 9.89 & 9.89 & 9.84 & 5.05 & 2.23 & 5.22 & 0.84 & 0.84 & 0.79 \\
VE & True & 2.21 & 1.70 & \textbf{2.50} & 9.90 & 9.92 & \textbf{9.89} & \textbf{4.99} & 2.21 & 5.19 & 4.56 & 3.97 & 4.33 \\
\hline
\end{tabular}
}
\caption{Averaged measures for the transonic cylinder dataset. $\downarrow$ indicates lower is better; $\uparrow$ indicates higher is better.}
\label{tab:CylinderMetrics}
\end{table}

\subsection{TurbRad}
\label{par:turb rad results}

\begin{figure}[t]
\centering
\includegraphics[width=\linewidth,keepaspectratio=True]{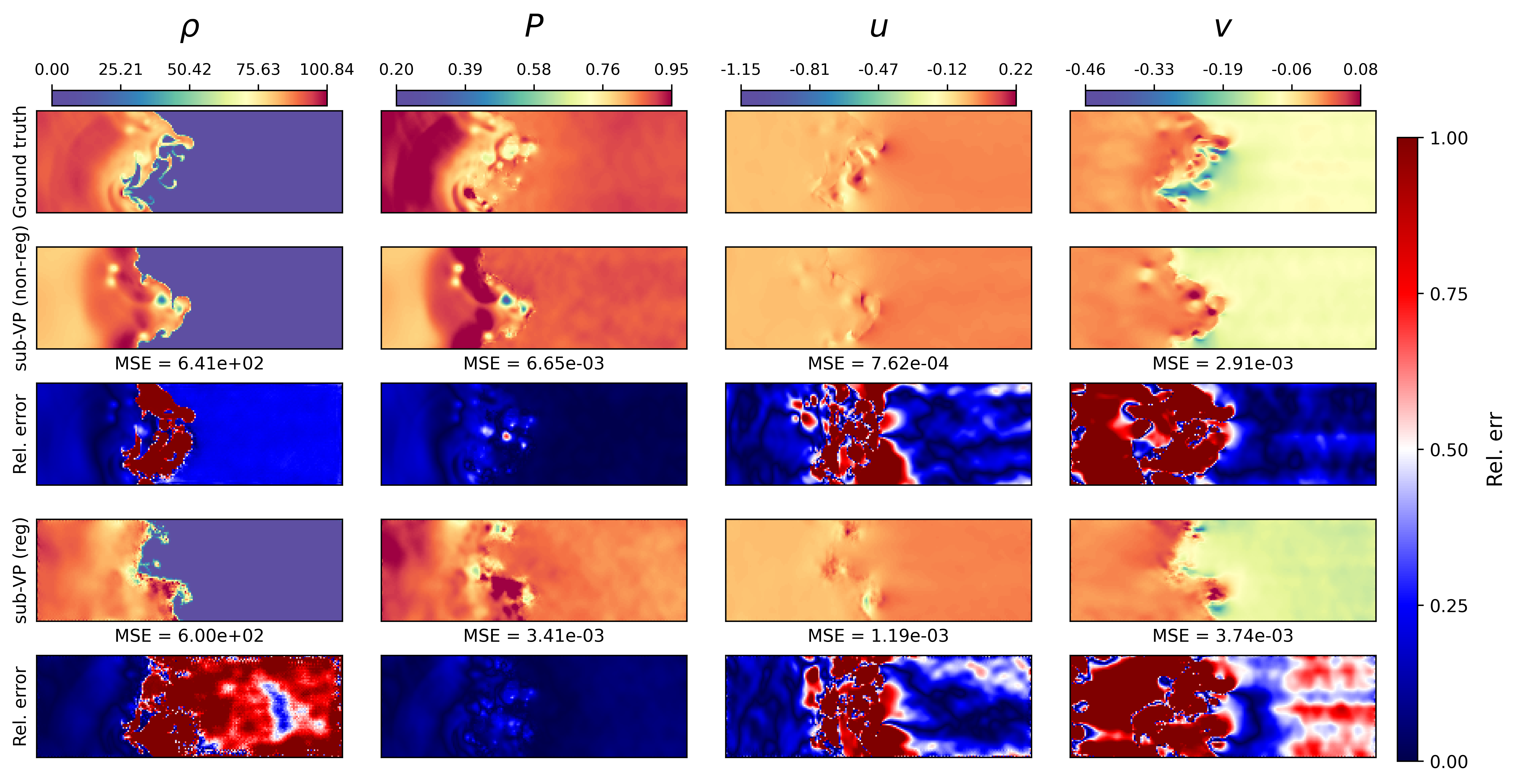}
\caption{Example test sample comparison at time $\tau = 60$ across fields between a sub-VP SDE-trained model without (on-reg) and with (reg) regularization for TurbRad at $\tau_\text{cool}=0.06$ (test case 1). The mean squared error (MSE) and relative error (Rel. error) are also shown.}
\label{fig:turbRadPred}
\end{figure}

\begin{figure}[t]
\centering
    \includegraphics[width=\linewidth]{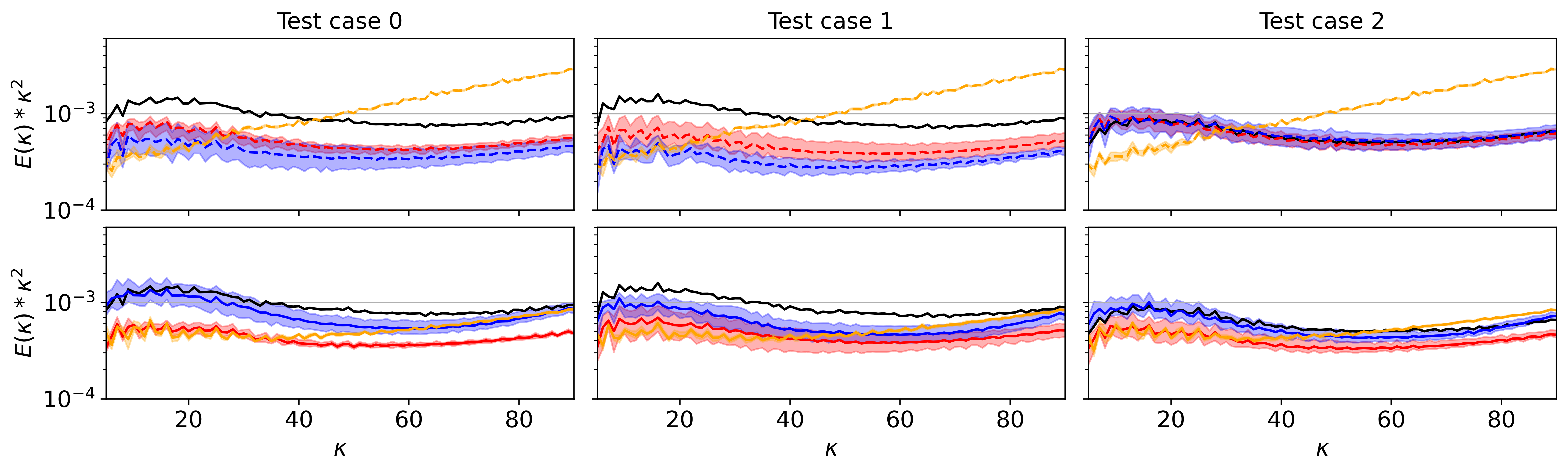}
        \vspace{-1em}
\caption{Time averaged spatial turbulent kinetic energy (spectra) of non-regularized (top, dashed lines) and regularized (bottom, solid lines) trained models for the TurbRad dataset. Shaded area shows the 5th to 95th percentile across sampled trajectories. \textcolor{black}{\rule[1.5pt]{0.5cm}{1pt}} ground-truth; \textcolor{red}{\rule[1.5pt]{0.5cm}{1pt}} VP SDE; \textcolor{blue}{\rule[1.5pt]{0.5cm}{1pt}} sub-VP SDE; \textcolor{orange}{\rule[1.5pt]{0.5cm}{1pt}} VE SDE.
}
\label{fig:turbRadSpectra}
\end{figure}

The TurbRad dataset presents a substantially more demanding setting than the transonic cylinder case, primarily due to its inherent multiscale complexity and the presence of sharp, discontinuity-like features in the density field $\density$, which often resemble membrane-like interfaces. These characteristics introduce significant challenges for both training stability and predictive accuracy. In practice, convergence is markedly slower in this case: while moderate stabilization is observed around epoch 500, full convergence remains elusive, and training is terminated at epoch 700 due to diminishing returns.

As shown in \fref{fig:turbRadSpectra}, the reconstruction of the energy spectrum highlights considerable discrepancies between SDE variants. In contrast to the cylinder configuration, where the VP SDE consistently outperformed others across all metrics, no such clear dominance is observed for the TurbRad scenario. Nevertheless, the sub-VP SDE demonstrates an enhanced spectral accuracy (see \tref{tab:TurbRadMetrics}), particularly in the high-frequency regime, suggesting an improved capacity to resolve fine-scale structures when the regularization is applied. Consequently, this formulation is retained for the final-time visual comparison shown in \fref{fig:turbRadPred}. In this setup, it is shown that the spectral advantage comes with a notable trade-off. The sub-VP SDE tends to amplify prediction noise across most physical quantities, with the exception of the pressure field $\Pres$, which appears robust to the induced variability. The VE SDE, in particular, displays highly unstable behavior without regularization, frequently producing noisy and incoherent predictions. This demonstrates its sensitivity to hyperparameter selection and the need for finer tuning. Regularization proves essential in mitigating this instability: it not only improves spectral accuracy, as evidenced by reduced divergence in the high-frequency range (see \fref{fig:turbRadSpectra}), but also leads to significant visual improvements. These gains are most clearly observed in the density field $\density$, where fine gradient structures on the left side are recovered, and high-frequency noise is suppressed, particularly for the VE formulation, as shown in \fref{fig:TurbRad_time_series} in \ref{app:time_series}.

Quantitative results in \tref{tab:TurbRadMetrics} indicate a general decline in performance compared to simpler cases, largely due to high MSE values driven by significant variance in the density field $\density$ and its wide dynamic range. Nonetheless, the evolution of the velocity magnitude $\norm{\uv}$ throughout the sequence (shown in \fref{fig:U_mag} in \ref{app:general_compa}) illustrates the qualitative benefits of regularization. The regularized sub-VP SDE achieves improved fidelity, accurately capturing both the structure and amplitude of the reference field over time. Interestingly, while the regularized variant offers the best visual match, the unregularized VP SDE still performs competitively, particularly in reproducing the velocity magnitude, underscoring the intrinsic robustness of the VP formulation and suggesting that regularization, though helpful, is not always essential for acceptable results. The autoregressive architecture continues to produce temporally consistent outputs, and key flow structures remain intact. As shown in \fref{fig:turbRadPred}, regularization consistently enhances the ability of all SDE variants to resolve fine-scale spatial features under challenging conditions. While no single formulation dominates across all metrics, the findings demonstrate that score-based diffusion models can robustly handle chaotic, multiscale flows, provided that regularization and tuning are carefully applied.

\begin{table}[t]
\centering
{
\begin{tabular}{cc ccc ccc ccc ccc}
\hline
\textbf{SDE} & \textbf{Reg.} & \multicolumn{3}{c}{$\taverage{\MSE} \downarrow$ ($10^{2}$)} & \multicolumn{3}{c}{$\taverage{\pearson} \uparrow$ ($10^{-1}$)} & \multicolumn{3}{c}{$\taverage{\DKL} \downarrow$ ($10^{1}$)} & \multicolumn{3}{c}{$\log-\MSE \downarrow$ ($10^{-1}$)} \\ \cline{3-14} 
& & tc 0 & tc 1 & tc 2 & tc 0 & tc 1 & tc 2 & tc 0 & tc 1 & tc 2 & tc 0 & tc 1 & tc 2 \\ \hline
VP & False & \textbf{0.66} & 0.77 & 0.69 & \textbf{9.52} & 9.45 & 9.50 & 1.05 & 1.11 & 0.90 & 0.67 & 0.99 & \textbf{0.04} \\
sub‑VP & False & 0.74 & 1.00 & \textbf{0.58} & 9.48 & 9.30 & \textbf{9.59} & 1.24 & 1.32 & 0.93 & 1.50 & 2.06 & 0.06 \\
VE & False & 0.96 & 1.02 & 0.97 & 9.32 & 9.28 & 9.32 & \textbf{0.61} & \textbf{0.50} & \textbf{0.37} & 1.40 & 1.49 & 1.77 \\
\hline
VP & True & 0.72 & \textbf{0.76} & 0.77 & 9.48 & \textbf{9.46} & 9.46 & 1.20 & 1.05 & 0.92 & 1.18 & 1.02 & 0.37 \\
sub‑VP & True & 0.93 & 0.85 & 0.78 & 9.32 & 9.40 & 9.45 & 0.91 & 0.91 & 0.78 & \textbf{0.17} & \textbf{0.38} & 0.06 \\
VE & True & 0.91 & 0.93 & 0.73 & 9.38 & 9.38 & 9.50 & 0.85 & 0.77 & 0.74 & 0.79 & 0.82 & 0.20 \\
\hline
\end{tabular}
}
\caption{Averaged measures for the TurbRad dataset. $\downarrow$ indicates lower is better; $\uparrow$ indicates higher is better.}
\label{tab:TurbRadMetrics}
\end{table}

\subsection{MHD}
\label{par:mhd results}

\begin{figure}[t]
\centering
\includegraphics[width=\linewidth,keepaspectratio=True]{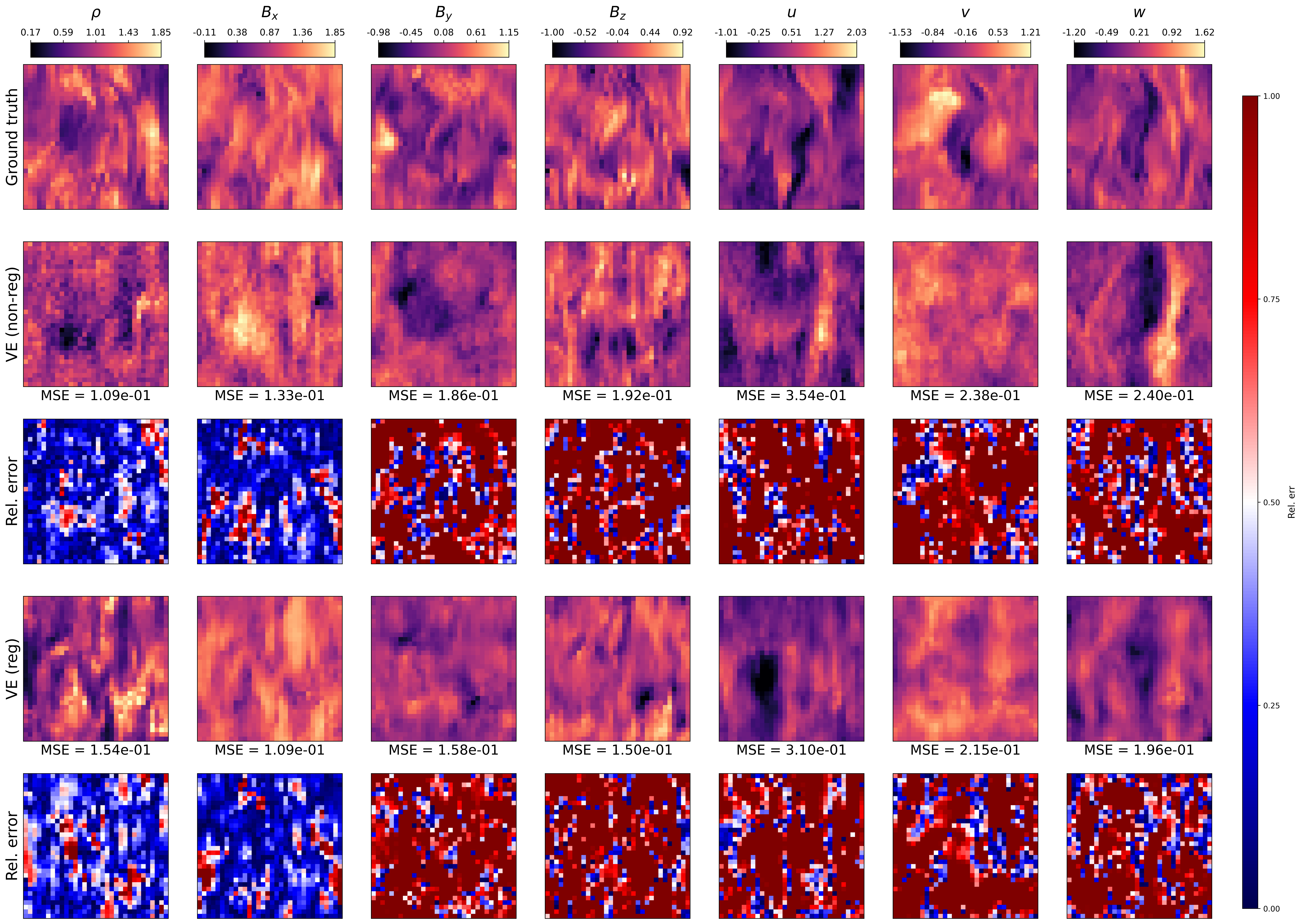}
\caption{Example test sample comparison at time $\tau = 20$ across fields between a VE SDE-trained model without (non-reg) and with (reg) regularization for MHD at $\Mach_s=0.7$ (test case 1) and position $z=15$. The mean squared error (MSE) and relative error (Rel. error) are also shown.}
\label{fig:MHD_pred}
\end{figure}

\begin{figure}[t]
\centering
    \includegraphics[width=\linewidth]{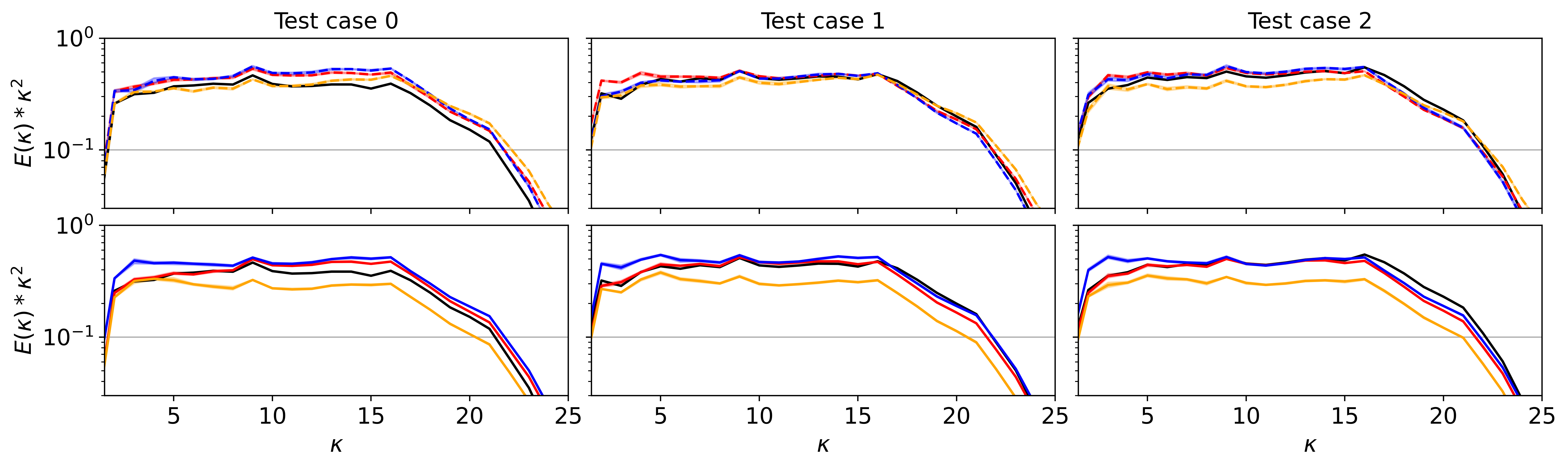}
    \vspace{-1em}
\caption{Time averaged spatial turbulent kinetic energy (spectra) of non-regularized (top, dashed lines) and regularized (bottom, solid lines) trained models for the MHD dataset. \textcolor{black}{\rule[1.5pt]{0.5cm}{1pt}} ground-truth; \textcolor{red}{\rule[1.5pt]{0.5cm}{1pt}} VP SDE; \textcolor{blue}{\rule[1.5pt]{0.5cm}{1pt}} sub-VP SDE; \textcolor{orange}{\rule[1.5pt]{0.5cm}{1pt}} VE SDE.
}
\label{fig:MHD_spectrum}
\end{figure}


The MHD case introduces significantly more complex dynamics compared to the transonic cylinder case, as evidenced by the high variability and multi-scale structure of the flow field. Due to the computational cost and instability of training in 3D, the number of training epochs is capped at 70. Despite this limitation, convergence is reached around 40 epochs, with a high relative loss, proving the difficulty of optimization in this setting. The volumetric nature of the dataset, coupled with strong gradients and intricate couplings between magnetic and velocity fields, makes this case a challenging benchmark for generative modeling.

At first glance, the spectral analysis in \fref{fig:MHD_spectrum} indicates no overall better formulation among the SDE formulations. Nevertheless, a few key trends emerge. The VE SDE, when unregularized, appears to yield the most accurate energy spectrum. However, a closer look at the late-time predictions in \fref{fig:MHD_pred} reveals the presence of residual noise, which undermines the credibility of its spectral accuracy. This discrepancy highlights the limitations of relying solely on spectral fidelity to assess generative performance in turbulent 3D flows. In contrast, the regularized VE SDE demonstrates substantial noise reduction and improved structure preservation, particularly on the $B_y$ component at slice $z = 15$, as shown in the time-series of \fref{fig:MHD_time_series} in \ref{app:time_series}. This confirms that regularization can partially compensate for the inherent instability of the VE formulation.

Temporal evolution analysis in \fref{fig:MHD_pred} shows that, despite the inherent difficulty of the task, the diffusion models are capable of reproducing coherent magnetohydrodynamics structures in the early phases of prediction. From $\tau = 5$ to $\tau = 10$, we can observe consistent morphological features across models for the magnetic field. These signatures suggest that the models can briefly retain fidelity to the coupled evolution of $\uv$ and $\bft{B}$. Beyond this range, however, the predictions begin to diverge significantly from the ground truth. The degradation is not purely random but rather manifests as a progressive distortion of the field topology, where magnetic tension and pressure forces are no longer properly balanced.

Among all models, the regularized VP SDE variant shows the most balanced performance: its spectral reconstruction is robust, and, most notably, it delivers substantial improvements in the velocity magnitude field $\norm{\uv}$, as shown in \fref{fig:U_mag} in \ref{app:general_compa}. This improvement is an encouraging sign of the regularization's effectiveness in enhancing large-scale dynamics. However, even in this case, performance degradation is observed in terms of statistical metrics. As reported in \tref{tab:MHDMetrics}, the $\MSE$ remains high, primarily due to strong density fluctuations and extreme value disparities across the field. Furthermore, a general decline in the $\pearson$ indicates growing temporal and spatial divergence from the ground truth, consistent with the challenges posed by the emergence of true turbulence-like structures and autoregressive prediction over long horizons. Despite these limitations, all models still succeed in reproducing coherent prediction and capturing multi-scale variability in the predicted fields. The chosen conditioning strategy and model architecture remain effective in this context, with spectral reconstruction remaining satisfactory across most frequencies. To better assess the long-term behavior and coherence, we include in \fref{fig:MHD_pred} the late-time predictions of each model, providing complementary qualitative insight beyond spectral metrics. This visualization is critical in 3D cases, where coherent spatial structure serves as the best practical evidence of model reliability, an aspect that is supported by the 3D rendering in \fref{fig:VP_U_mag} and \fref{fig:VP_B_mag} in \ref{app:time_series}.

In summary, while the MHD case exhibits an overall decrease in predictive performance due to the inherent complexity of 3D MHD turbulence, diffusion models, especially the regularized VP SDE, remain capable of delivering meaningful reconstructions with the proper order of magnitude. The results confirm that regularization enhances stability, spectral fidelity, and field coherence, and remains a critical component for extending diffusion-based generative approaches to high-dimensional physical systems.

\begin{table}[t]
\centering
{
\begin{tabular}{cc ccc ccc ccc ccc}
\hline
\textbf{SDE} & \textbf{Reg.} & \multicolumn{3}{c}{$\taverage{\MSE} \downarrow$ ($10^{-1}$)} & \multicolumn{3}{c}{$\taverage{\pearson} \uparrow$ ($10^{-1}$)} & \multicolumn{3}{c}{$\taverage{\DKL} \downarrow$ ($10^{1}$)} & \multicolumn{3}{c}{$\log-\MSE \downarrow$ ($10^{-2}$)} \\ \cline{3-14} 
& & tc 0 & tc 1 & tc 2 & tc 0 & tc 1 & tc 2 & tc 0 & tc 1 & tc 2 & tc 0 & tc 1 & tc 2 \\ \hline
VP & False & 1.43 & 1.39 & 1.79 & 7.85 & 8.05 & 7.65 & 0.94 & 0.78 & 0.85 & 1.17 & 0.38 & 0.32 \\
sub‑VP & False & 1.50 & 1.21 & 1.59 & 7.70 & 8.19 & 7.78 & 0.94 & \textbf{0.75} & 0.89 & 1.20 & \textbf{0.15} & \textbf{0.28} \\
VE & False & \textbf{1.23} & 1.11 & \textbf{1.38} & \textbf{8.06} & 8.28 & \textbf{7.98} & 0.91 & 0.86 & 0.86 & 1.40 & 0.40 & 0.50 \\
\hline
VP & True & 1.38 & 1.23 & 1.62 & 7.85 & 8.13 & 7.77 & \textbf{0.87} & 0.80 & \textbf{0.80} & \textbf{0.44} & 0.24 & 0.47 \\
sub‑VP & True & 1.47 & 1.45 & 1.68 & 7.82 & 7.99 & 7.76 & 0.92 & 0.76 & 0.84 & 1.78 & 0.44 & 0.51 \\
VE & True & 1.24 & \textbf{1.05} & 1.40 & 8.02 & \textbf{8.35} & 7.97 & 1.00 & 0.82 & 0.87 & 1.44 & 3.17 & 4.05 \\
\hline
\end{tabular}
}
\caption{Averaged measures for the MHD dataset. $\downarrow$ indicates lower is better; $\uparrow$ indicates higher is better.}
\label{tab:MHDMetrics}
\end{table}

\subsection{Extension to filtering}

\subsubsection{Theoretical implementation}

A significant challenge associated with the regularization technique is its inherent sensitivity to hyperparameter tuning. Achieving an appropriate balance is crucial; otherwise, residual noise can remain in the output images, adversely affecting the prediction quality. This phenomenon is particularly apparent in the density field prediction illustrated in \fref{fig:turbRadPred}. Due to the field's intrinsic high-scale variance, any residual noise is significantly amplified, making these artifacts a major contributor to the overall prediction error. To highlight this issue one may be examining derived quantities such as the vorticity $\vorticity=\bnabla\times\uv$ and the $Q$-criterion \cite{DUB00}:
\begin{equation}\label{eq:critQ}
Q=\demi(\bft{\Omega}:\bft{\Omega}-\bft{D}:\bft{D})\,,
\end{equation}
where $\bft{D}$ is the symmetric velocity rate tensor \eqref{eq:strainvelocity}, $\bft{\Omega}=\smash{\demi(\bnabla\otimes\uv-(\bnabla\otimes\uv)^\itr)}$ is the skew-symmetric rotation rate tensor, and $\bft{A}:\bft{B}=\smash{\sum_{i,j}A_{ij}B_{ij}}$. In 2D ($d=2$) the vorticity is actually a scalar field $\vortj=\curl\cdot\uv=-\partial_y u+\partial_x v$ where $\curl\equiv(-\partial_y,\partial_x)^\itr$, and:
\begin{equation*}
\bft{D} = \demi
\begin{bmatrix}
2 \partial_x u & \partial_y u + \partial_x v \\
\partial_y u + \partial_x v & 2 \partial_y v
\end{bmatrix}
\,, \quad
\bft{\Omega} = \demi
\begin{bmatrix}
0 & -\vortj \\
\vortj & 0
\end{bmatrix}
\,.
\end{equation*}
The $Q$-criterion balances between the rotation rate and the strain rate and its positive iso-surfaces (or iso-values in 2D) display areas (or lines) where the strength of rotation overcomes the strain, highlighting the vortex envelopes; see \fref{fig:iso_false} and \fref{fig:iso_true}.

Nonetheless, since the regularization already preserves the overall structural integrity effectively, we propose complementing it with an adaptive filtering technique, such as a Perona-Malik filter \cite{PER90}, to further mitigate residuals without compromising the main flow features. This filter employs anisotropic diffusion, selectively smoothing homogeneous regions while retaining sharp gradients like edges or coherent structures. The mathematical formulation of the Perona-Malik anisotropic diffusion equation is expressed as:
\begin{equation}
    \partial_{t}\bft{i} = \bdiv{_\rv} \parenthesis{D \parenthesis{\left \Vert \bnabla_\rv{\bft{i}} \right \Vert} \bnabla_\rv \bft{i}} \,,
\end{equation}
where $\bft{i}$ is the image to be processed (\emph{e.g.}, the velocity field $\uv$) and $D$ is a diffusion coefficient. Physically, this filter mimics a controlled diffusion process where diffusion is inhibited across strong gradients and enhanced in flatter regions. As a result, small-scale noise and high-frequency fluctuations are suppressed, while the larger, physically meaningful structures such as vortices, shear layers, and shock-like interfaces are preserved. This makes the Perona-Malik filter particularly suited to fluid dynamics data, where maintaining the integrity of coherent structures is essential for accurate analyses. The discrete implementation is given in~\aref{algo}.

\begin{algorithm}[H] 
\caption{\hspace{-.2em}: Discrete Perona-Malik filtering on velocity field} \label{algo}
\begin{algorithmic}[1]

\State Set $\uv(\rv, \tau) = \uv^{(0)}$, $\gamma=0.05$, $\eta=0.03$ and $\varepsilon = 10^{-8}$.

\For{$n = 1$ \textbf{to} $N$}

    \State Set $\bft{G}=(G_x, G_y)^{\top}$, where $G_x = \demi\parenthesis{\uv^{(n)}_{i+1,j} - \uv^{(n)}_{i-1,j}}$ and $G_y = \demi\parenthesis{\uv^{(n)}_{i,j+1} - \uv^{(n)}_{i,j-1}}$.
    \State Compute $D = \parenthesis{1 + \left(\left \Vert \bnabla\uv \right \Vert/\gamma\right)^2}^{-1}$ with $\left \Vert \bnabla\uv \right \Vert = \sqrt{G_x^2 + G_y^2 + \varepsilon}$.
    \State Set $\bft{J} = D \cdot \bft{G} =(J_x, J_y)^{\top}$.
    \State Update $\uv^{(n+1)} = \uv^{(n)} + \eta \, \bdiv{} \bft{J}$, where $\bdiv{} \bft{J} = \demi\parenthesis{J_{x,i+1,j} - J_{x,i-1,j}} + \demi\parenthesis{J_{y,i,j+1} - J_{y,i,j-1}}$.
    
\EndFor
\State \textbf{return} $\hat{\uv}=\uv^{(N)}$

\end{algorithmic}
\end{algorithm}
Note that in~\aref{algo}, although the velocity field $\uv$ is tensor-valued, the filtering is applied componentwise. Each scalar component of $\bft{u}$ is treated as a 2D image over the spatial grid, and the nonlinear diffusion process is performed independently on each. The operator $\bdiv{}\bft{J}$ denotes a finite-difference approximation of the divergence, applied separately to each scalar flux field $\bft{J}$ associated with a single component of $\bft{u}$. This treatment follows established practices in image processing and vector field regularization. The threshold parameter $\gamma$, also referred to as the diffusion or edge threshold, is set to a low value to separate noise from meaningful structures such as sharp features or interfaces in the flow field, while preserving high-frequency components.
To assess the effect of the filtering technique, we propose to apply it on the best scenario in \sref{algo_app} below, namely the transonic cylinder case, since this case exhibits the best performances for the regularization technique and the most coherent structures overall.

\subsubsection{Structural improvements} \label{algo_app}

\begin{figure}[t]
    \centering
    \includegraphics[width=.95\textwidth]{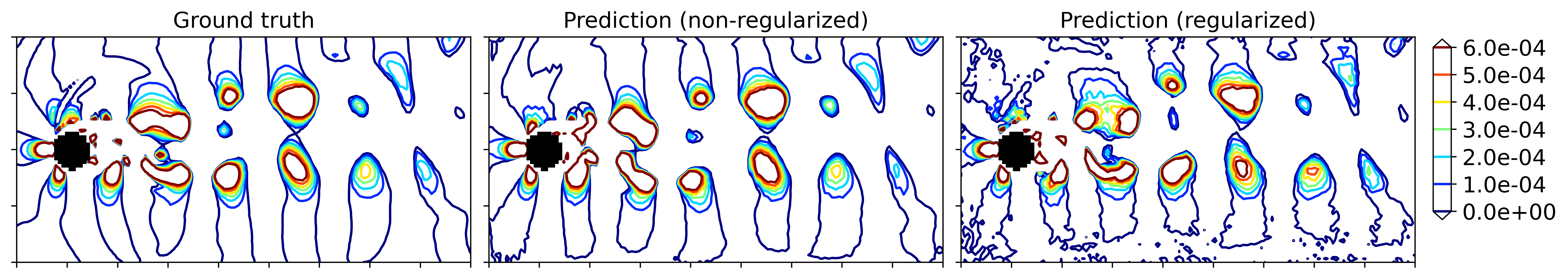}
    \includegraphics[width=.95\textwidth]{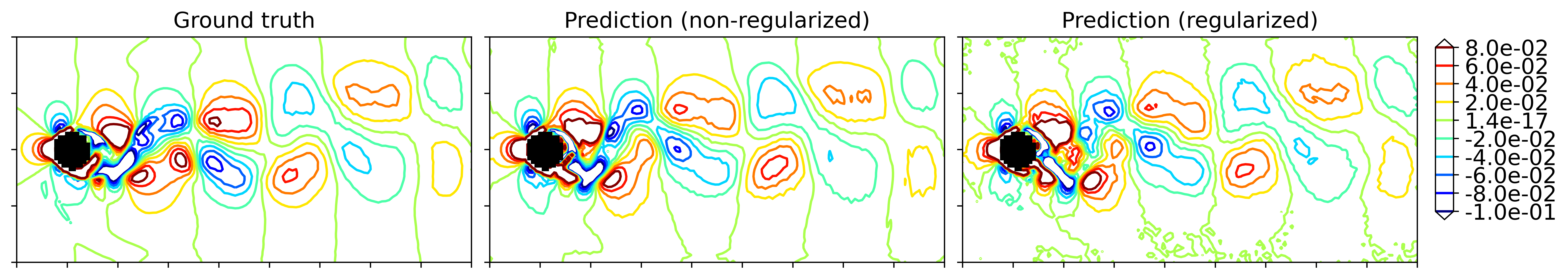}
    \caption{Iso-values of positive $Q$-criterion (top row) and vorticity field (bottom row) for test case 1 for the VP SDE trained model with the transonic cylinder dataset, time $\tau=30$.}
    \label{fig:iso_false}

    \vspace{1em}

    \centering
    \includegraphics[width=.95\textwidth]{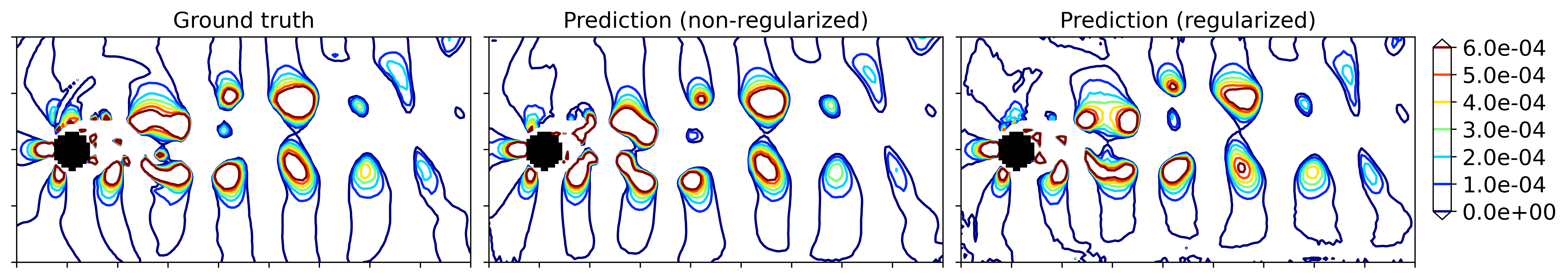}
    \includegraphics[width=.95\textwidth]{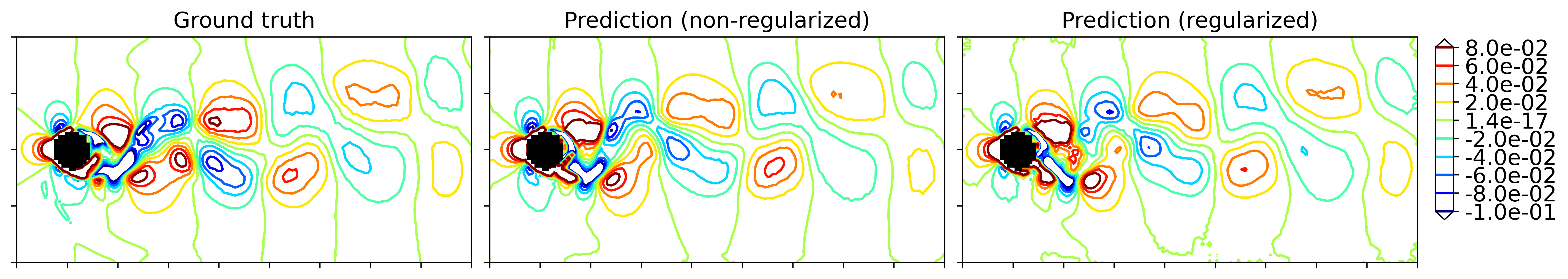}
    \caption{Iso-values of positive $Q$-criterion (top row) and vorticity field (bottom row) for test case 1 for the VP SDE trained model with the transonic cylinder dataset after filtration, time $\tau=30$.}
    \label{fig:iso_true}
    
\end{figure}

\fref{fig:iso_false} and \fref{fig:iso_true} present a qualitative comparison of model predictions with ground truth data from the transonic cylinder dataset at simulation step $\tau = 30$, using two flow diagnostics: positive iso-values of the $Q$-criterion (top row) and the vorticity field (bottom row). Each figure is arranged as a 2 by 3 grid with columns representing the ground truth, the non-regularized and regularized predictions from a VP model. The $Q$-criterion identifies coherent vortex structures by assessing the local balance between strain and rotation rates, while the vorticity field captures rotational features through the curl of velocity.

\fref{fig:iso_false} displays raw predictions without post-processing. The ground truth shows a well-defined staggered vortex street, with coherent, spatially ordered $Q$-structures downstream of the cylinder. The non-regularized model replicates most of these features, although with some distortion in shape and amplitude farther downstream. The regularized model, however, introduces visible noise and fine-scale artifacts, disrupting the continuity and clarity of the vortex structures, especially near the cylinder, and on the outer boundaries. This suggests that the regularization, while beneficial for generalization or convergence, may promote high-frequency noise incompatible with physical flow features. Vorticity fields show similar trends. Ground truth vorticity reveals a clean alternation of positive and negative lobes characteristic of vortex shedding. The non-regularized model captures this pattern with reasonable fidelity, despite some amplitude and diffusion mismatches. In contrast, the regularized version yields fragmented, disorganized vorticity patches, especially in the near-cylinder region, implying that the regularization perturbs spatial coherence critical for representing flow dynamics.

\fref{fig:iso_true} presents the same diagnostics after applying a Perona–Malik anisotropic diffusion filter to the velocity field. As previously discussed, this filtering technique attenuates high noise components while preserving critical gradients. Post-filtering, both the regularized and non-regularized predictions of the $Q$-criterion and vorticity fields appear significantly cleaner and more coherent, with vortex structures and wake patterns more closely resembling the ground truth. These results suggest that the model captures key physical features, requiring only minimal \emph{post hoc} denoising.

These findings highlight an inherent trade-off: while regularization tends to improve numerical stability and training convergence, it may introduce localized noise artifacts. In contrast, non-regularized models, though potentially less stable over long horizons, tend to produce smoother predictions. The effectiveness of the Perona–Malik filter in enhancing visual and structural consistency emphasizes the role of post-processing when evaluating generative models for fluid dynamics, and it further motivates the development of evaluation metrics that assess physical consistency beyond simple reconstruction error.

\section{Conclusion}\label{sec:conclusion}

Across a range of increasingly complex fluid dynamics scenarios, namely, the transonic flow, turbulent radiative layer, and magnetohydrodynamics, the regularized VP SDE emerges as the most consistently robust and versatile formulation among the score-based diffusion models evaluated. It demonstrates superior performance in both spectral reconstruction and physical field fidelity, particularly for simpler test cases such as the transonic cylinder flow, where it outperforms all alternatives across quantitative metrics and qualitative diagnostics. Its temporal stability, and spectral fidelity across both low and high-dimensional domains, makes it a strong candidate for scalable applications in data-driven fluid flow modeling. Regularization proves crucial in stabilizing training dynamics and preserving large-scale structures, offering significant advantages even in more turbulent or high-dimensional settings, such as the TurbRad and MHD datasets. Here, although no single model dominates across all criteria, regularized variants, especially the VP and sub-VP SDEs, exhibit marked improvements in coherence and high-frequency reconstruction. However, the benefits of regularization come with trade-offs. Notably, in the raw outputs, regularized models tend to introduce high-frequency noise, which compromises the clarity of physical features such as vortex structures and coherent shear layers. These artifacts are particularly visible with the $Q$-criterion and vorticity fields. Post-processing with anisotropic diffusion filtering, such as the Perona–Malik method, significantly mitigates these artifacts and recovers much of the physical realism present in the ground truth, confirming that the underlying generative mechanisms are capturing relevant dynamics. Overall, the findings support the viability of diffusion-based generative models for high-fidelity flow prediction tasks, provided that careful attention is paid to both model design (via appropriate SDE formulation and regularization) and post-processing. However, challenges remain, especially in long-horizon or highly multiscale regimes. Despite their promising performance, diffusion models remain computationally intensive and prone to divergence in long-horizon or highly nonlinear settings. While the present study highlights the effectiveness of diffusion models for flow prediction tasks, several challenges and opportunities remain for enhancing their applicability to complex physical systems. A central limitation lies in the sensitivity of these models to the conditioning strategy used during training and inference. Future work should aim to improve stability and scalability through more advanced conditioning methods \cite{shysheya2024conditional}, including physics-informed inputs, adaptive embeddings, or temporal attention mechanisms. Enhancing efficiency via multi-scale architectures, parameter sharing, or tailored noise schedules could also reduce training cost without sacrificing fidelity. Additionally, hybrid strategies that integrate diffusion models with classical solvers (using them as correctors, predictors, or data-driven priors) offer a promising path toward physically stable inference. Building on benchmarking insights from \cite{benchmarking}, future evaluations should expand to include constraints like covariance regularization as the one suggested in this paper. Nonetheless, it is worth emphasizing that our framework successfully achieved full-field generation in both 2D and 3D turbulent settings within reasonable time, demonstrating the practical feasibility of autoregressive diffusion models for spatiotemporal forecasting. This empirical result reinforces their potential as robust surrogates for complex flow evolution, even in high-dimensional regimes. Ultimately, while diffusion models open a new frontier for data-driven PDE solving, realizing their full potential will require targeted improvements in conditioning, efficiency, and physical constraint integration.

\clearpage  
\appendix
\section{Notations}
\label{Notations}



\subsection{Nested expectations}
\label{Nested expectations}
For clarity and conciseness, we introduce a compact notation to represent nested expectations over multiple random variables. Consider $n$ random variables $\bx_1, \dots, \bx_n$, where each $\bx_i$ is drawn independently from a corresponding distribution $p_i$. The expectation of a function $f(\bx_{1},\dots,\bx_{n})$ with respect to these variables can be denoted as:
\begin{equation}
\mathbb{E}_{
\begin{aligned}
& \scriptstyle \bx_{1} \sim p_1(\bx_{1}) \\[-1.em]
& \scriptscriptstyle \quad \,\, \vdots \\[-1.em]
& \scriptstyle \bx_{n} \sim p_n(\bx_{n})
\end{aligned}
}\!
\left\{
f(\bx_{1},\dots,\bx_{n})
\right\} :=
\esp_{
\bx_{1} \sim p_1(\bx_{1})} \cdots 
\esp_{
\bx_{n} \sim p_n(\bx_{n})}
\left\{
f(\bx_{1},\dots, \bx_{n})
\right\} \,.
\end{equation}
This notation compactly represents the sequence of expectations taken in order, and will be used throughout for readability when dealing with complex multi-variable expectations.

\subsection{Time indexing for simulations}
\label{Phi time notation}
Throughout this work, we denote by $\tau$ the physical simulation time, used as a superscript of the state variable $\bx(\bft{r},\tau)=\bx^\tau$ to indicate its value at time $\tau$. In practice, the simulation timeline is discretized into uniform or non-uniform intervals, leading to a slight abuse of notation. We use expressions such as $\partial_\tau$ to refer to continuous-time derivatives, while also using $\tau+1$ to designate the next discrete time index. It should be noted that $\tau+1$ does not correspond to $\tau+1$ in physical units (\emph{e.g.}, seconds), but simply refers to the subsequent index in the discretized simulation timeline. This slight overlap in notation is assumed to be clear from context.

\section{Finite-time Lyapunov exponents}
\label{sec:ftle_lyapunov_time}

    
    
    

The Finite-Time Lyapunov Exponent (FTLE) \cite{lya} is a widely used diagnosis for identifying regions of Lagrangian chaos in unsteady dynamical systems, particularly in fluid flows. It quantifies the rate of separation of infinitesimally close particles over a finite integration time. This section defines the FTLE formally, explores its connection with chaotic behavior, and introduces the notion of Lyapunov time.

\subsection{Definition of FTLE}

Let $\bs{\phi}_{\tau}(\rv_0,\tau_0) = \bs{\phi}_{\tau - \tau_0}(\rv_0,0)$ denote the flow map that brings a fluid material particle at initial position $\rv_0$ at time $\tau_0$ to its position at time $\tau$. Let $\tau = \tau_0 + \Tint$, where $\Tint$ is some finite integration lapse time. The strain-rate tensor of the flow map reads:
\begin{equation}
\bft{F}_{\Tint}(\rv_0) = \bnabla_{\rv_0} \bs{\phi}_{\Tint}(\rv_0,0) \, .
\end{equation}
The Cauchy–Green strain tensor is then defined as:
\begin{equation}
\bft{C}_{\Tint}(\rv_0) = (\bft{F}_{\Tint}(\rv_0))^\itr \bft{F}_{\Tint}(\rv_0)\,.
\end{equation}
Let $\smash{\lambda_\imax(\rv_0)}$ be the largest (positive) eigenvalue of $\smash{\bft{C}_{\Tint}(\rv_0)}$. Then the FTLE at position $\rv_0$ and time $\tau_0$ for $\Tint\neq0$ is defined as:
\begin{equation}
\chi_{\Tint}(\rv_0) = \frac{1}{\abs{\Tint}} \log \sqrt{\lambda_\imax(\rv_0)}\,.
\end{equation}
FTLE fields typically reveal sharp gradients and zones of trajectory divergence, which correspond to Lagrangian Coherent Structures (LCS), delimiting regions that govern the evolution of particles and mixing phenomena.

\subsection{FTLE and Lyapunov time}

A large FTLE reflects a strong exponential sensitivity to initial conditions over the finite-time interval $\Tint$. That is, even small perturbations in the initial position can lead to widely diverging trajectories. Unlike the classical (non-finite time) Lyapunov exponent, which describes asymptotic behavior as time tends to infinity, the FTLE provides a local (in time and space) characterization of this sensitivity. The associated Lyapunov time, denoted by $\TLya$, quantifies the typical time scale over which predictability is lost. If $\delta_0$ denotes an initial separation between two nearby trajectories, their separation after time $\tau$ evolves approximately as $\delta(\tau) \approx \smash{\delta_{0}\iexp^{\tau/\TLya}}$ with $\TLya$ verifying:
\begin{equation}
\TLya(\rv_0) = \frac{1}{\chi_{\Tint}(\rv_0)} \, .
\end{equation}
This quantity provides a local measure of predictability: the smaller the Lyapunov time, the faster the divergence of nearby trajectories. Consequently, regions of high FTLE correspond to small Lyapunov times and indicate zones of low predictability. Note that the value of $\Tint$ plays a crucial role in FTLE computation, a short $\Tint$ may fail to capture longer-time instabilities but a very long $\smash{\Tint}$ may cause trajectories to exit the domain or introduce numerical instability. In practice, $\smash{\Tint}$ is chosen such that FTLE gradients are sufficiently developed but still bounded within the physical domain. Here, the FTLE is computed over the entire temporal and spatial grid for a finite number of particles. The integration time $\smash{\Tint}$ is then heuristically adjusted to optimize trajectory separation.

\subsection{Global Lyapunov time and dataset application}

\begin{figure}[t]
    \centering
    \begin{minipage}[]{0.615\textwidth}
        \includegraphics[width=\linewidth]{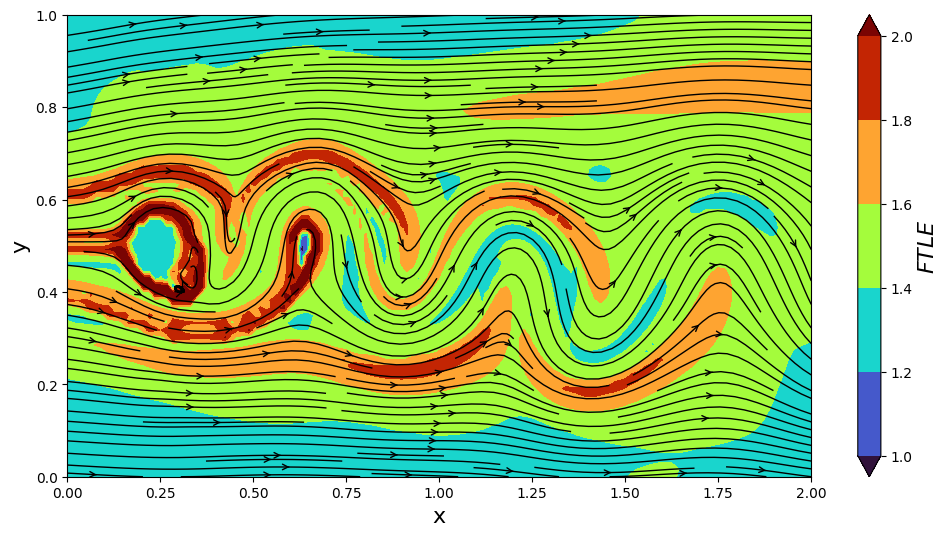}
    \end{minipage}
    \begin{minipage}[]{0.335\textwidth}
        \includegraphics[width=\linewidth]{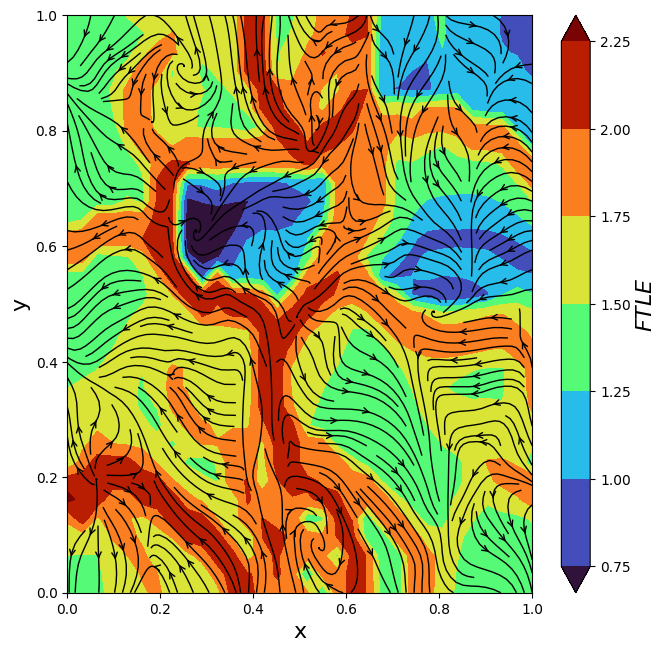}
    \end{minipage}

    \includegraphics[width=.725 \linewidth]{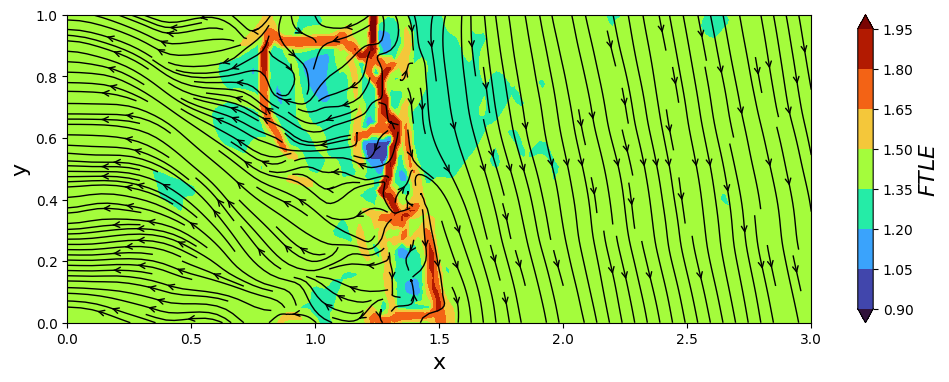}

    \caption{Streamlines and FTLE fields for different example flows. Streamlines indicate the instantaneous velocity direction, while the color contours show the FTLE. Top left: transonic flow; top right: MHD flow; bottom: TurbRad.}

    \label{fig:FTLE}
\end{figure}

In practice, a global Lyapunov time is defined to characterize the shortest time scale over which the system loses memory of its initial conditions due to exponential divergence. This time scale is typically estimated by taking the inverse of the maximum FTLE value over the spatial domain. By selecting $\smash{\TLya^\text{global}} = \smash{1/\max_{\rv_0}\chi_{\Tint}(\rv_0)}$, one captures the region of fastest separation, which effectively governs the breakdown of coherent flow structures. This strategy yields a conservative estimate of the predictability horizon, providing an upper bound on the system's sensitivity to initial conditions. This choice is useful when extracting  the stochastic fluctuations from a deterministic flow field as done in \sref{sec:regularization}. When the temporal mean is subtracted over a short interval $\Tint = 1$, one implicitly assumes that the flow decorrelates over this time scale. If the global Lyapunov time $\smash{\TLya^\text{global}} < 1$, this assumption is justified: the system has already experienced sufficient exponential divergence within one unit of time, allowing the residuals to be interpreted as effectively uncorrelated or random. We then compute the FTLE for each dataset, visualizations are adapted from the code by Richard Galvez\footnote{\href{https://github.com/richardagalvez/Vortices-Python/blob/master/Vortex-FTLE.ipynb}{https://github.com/richardagalvez/Vortices-Python/blob/master/Vortex-FTLE.ipynb}} (May 2015).
\begin{itemize}
\item For the Transonic cylinder and MHD datasets, the FTLE is computed with initial time $\tau_0 = 0$, as both flows exhibit periodic or quasi-periodic behavior.
\item For the TurbRad dataset, the initial time is shifted to $\tau_0 \approx 30$ to bypass the initial transient regime and capture chaotic dynamics.
\item Despite the MHD simulation being 3D, the FTLE is computed on a 2D plane. While this may lead to a slight underestimation of the true Lagrangian divergence, the results remain qualitatively consistent and visually informative.
\end{itemize}
Based on the FTLE fields, we estimate the global Lyapunov time $\TLya^\text{global}$ for each case (see \fref{fig:FTLE}):
\begin{itemize}
\item Transonic cylinder: $\TLya^\text{global} \approx 0.5 < 1$,
\item TurbRad: $\TLya^\text{global} \approx 0.51 < 1$,
\item MHD: $\TLya^\text{global} \approx 0.45 < 1$.
\end{itemize}

\section{Additional results}\label{app:add_results}

This section presents supplementary results extending the main content of the paper. \ref{app:general_compa} displays the temporal evolution of the overall velocity magnitudes, while \ref{app:time_series} provides visualizations of time series.

\subsection{General comparison} \label{app:general_compa}

\begin{figure}[t]
    \centering
\begin{minipage}[]{0.45\textwidth}
    \includegraphics[width=\linewidth]{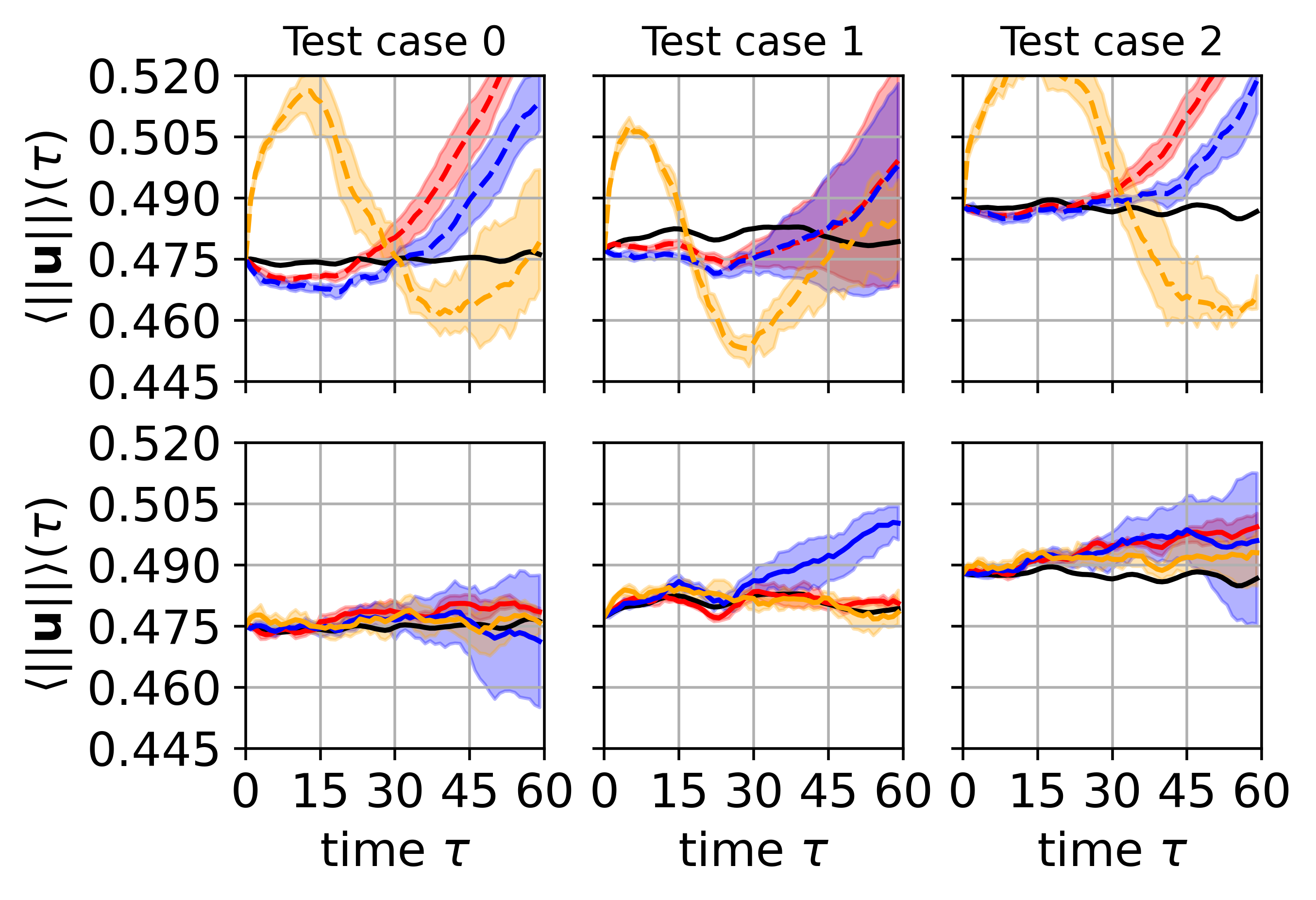}
\end{minipage}
\begin{minipage}[]{0.45\textwidth}
    \includegraphics[width=\linewidth]{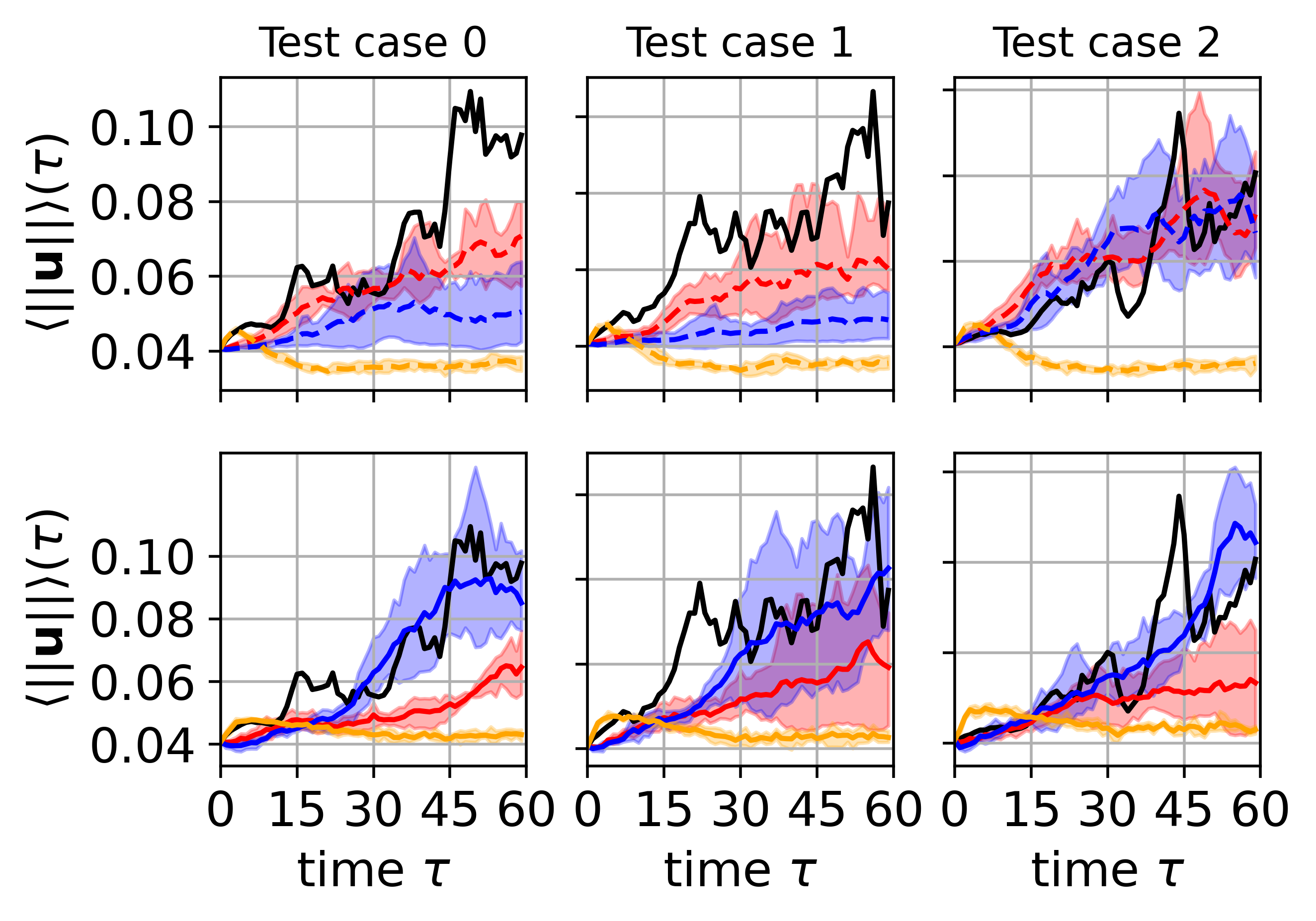}
\end{minipage}
\begin{minipage}[]{0.45\textwidth}
    \includegraphics[width=\linewidth,keepaspectratio=true]{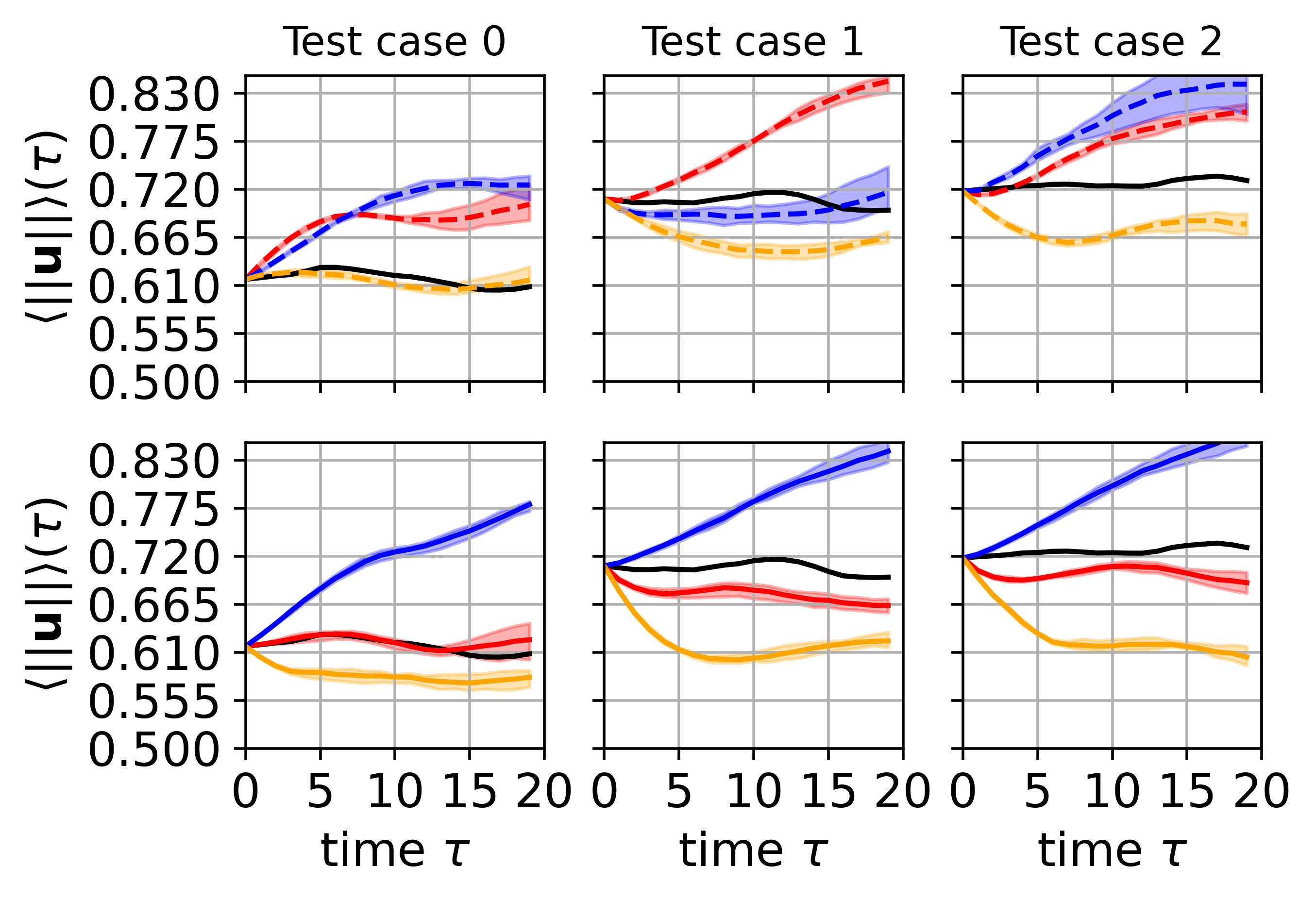}
\end{minipage}
    \caption{Temporal evolution of the spatially averaged velocity magnitude $\saverage{\norm{\uv}}$ for the transonic cylinder (top-left), TurbRad (top-right), and MHD (bottom) datasets. These plots highlight the global dynamical activity and temporal variability of the flows across datasets. Shaded areas show the 5th to 95th percentile across sampled trajectories. For each group, non-regularized trained models are top, dashed line graphs, regularized trained models are bottom, solid line graphs. \textcolor{black}{\rule[1.5pt]{0.5cm}{1pt}} ground-truth; \textcolor{red}{\rule[1.5pt]{0.5cm}{1pt}} VP SDE; \textcolor{blue}{\rule[1.5pt]{0.5cm}{1pt}} sub-VP SDE; \textcolor{orange}{\rule[1.5pt]{0.5cm}{1pt}} VE SDE.
    }
    \label{fig:U_mag}
\end{figure}

\fref{fig:U_mag} illustrates the temporal evolution of the spatially averaged velocity magnitude $\tau\to\saverage{\norm{\uv}}(\tau)$. It highlights the global dynamics and temporal variability of the flow across different regimes. In the transonic cylinder configuration, the dynamics appear oscillatory or weakly varying. The VP and sub-VP SDEs closely follow the reference trajectory during early time steps, with the regularized variants maintaining this alignment further in time. While most models remain in reasonable agreement with the reference, the VE SDE stands out by exhibiting larger deviations and broader uncertainty bands. Regularization clearly mitigates this effect, as seen from the reduced spread in the shaded confidence regions. In the TurbRad dataset, pronounced temporal variability is observed. Here, the regularized VP and sub-VP SDEs perform slightly better in tracking the reference evolution while maintaining tighter confidence bounds. Among them, the regularized sub-VP achieves the closest match overall. The MHD dataset, on the other hand, features a steadily increasing velocity magnitude. In this case, the regularized sub-VP model shows a consistent overshoot but retains coherent trends, whereas the VE SDE persistently underestimates the flow magnitude. The regularized VP SDE yields the most stable results overall. Across all datasets, regularization enhances model stability for several SDE cases and reduces variance, especially for the VP-based formulations, though the effectiveness varies with the specific flow characteristics.

\subsection{Time series} \label{app:time_series}

\begin{figure}[t]
    \centering
    \includegraphics[width=\textwidth]{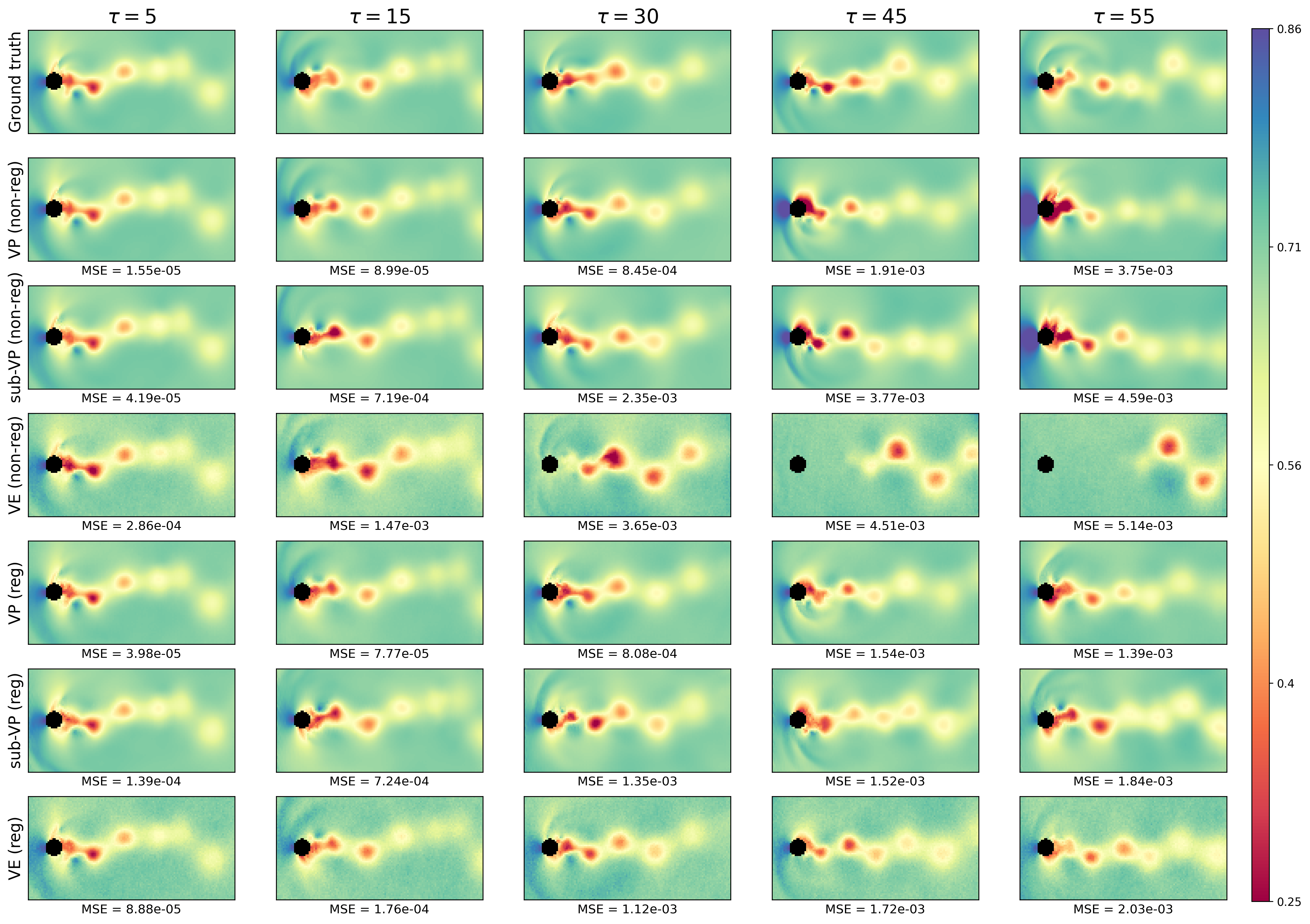}
    \vspace{-1em}
    \caption{Time series of the pressure field $\Pres$ for test case 2 of the transonic cylinder dataset.}
    \label{fig:Cylinder_time_series}

    \vspace{1.5em}

    \centering
    \includegraphics[width=\textwidth]{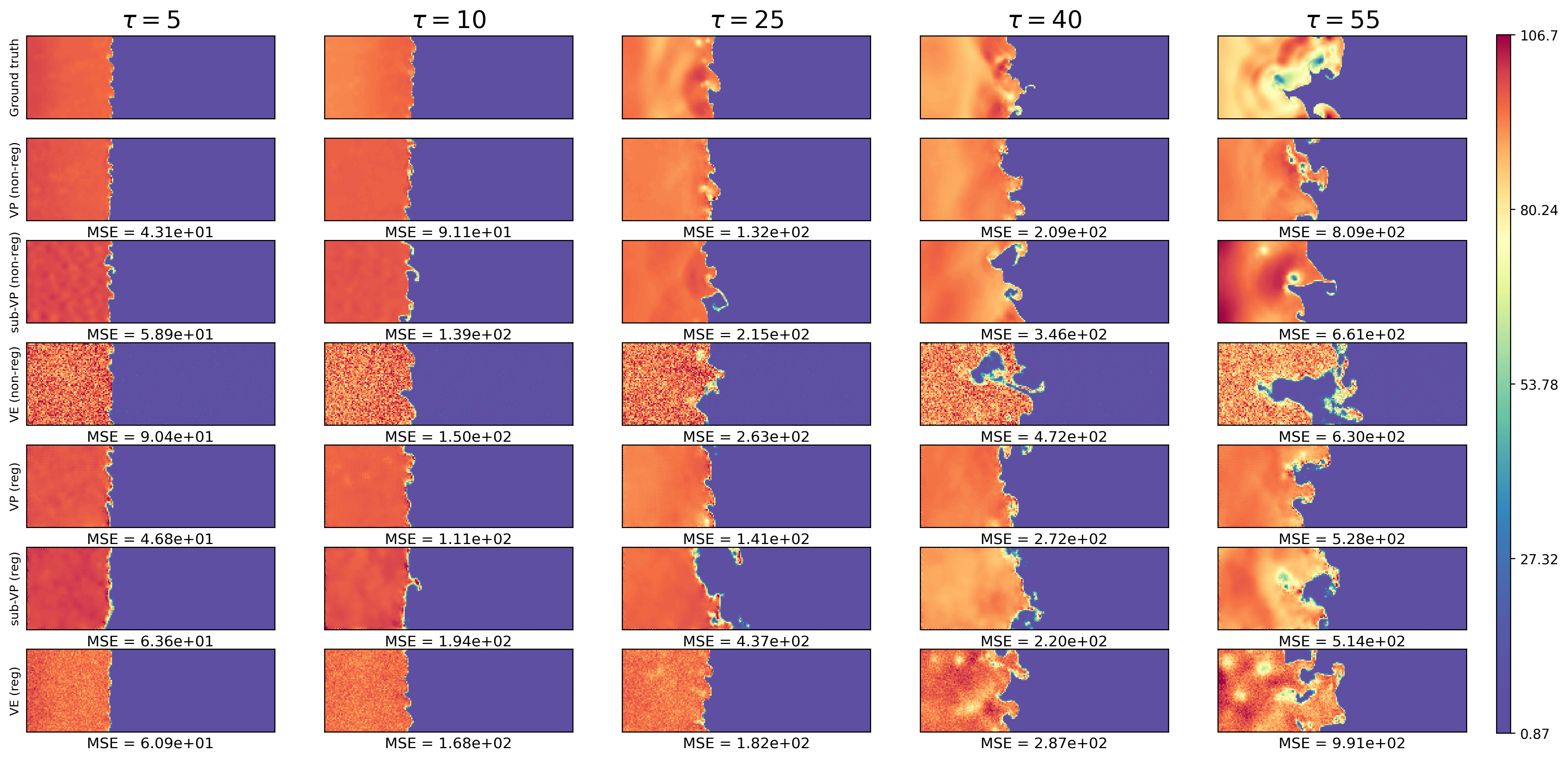}
    \vspace{-1em}
    \caption{Time series of the density field $\density$ for test case 2 of the TurbRad dataset.}
    \label{fig:TurbRad_time_series}
\end{figure}

We aim to present the most discriminative visualizations using time series at various time offsets $\tau$, based on representative combinations of test cases and physical fields. This selection is made without introducing preferential bias, but rather prioritizes the clarity and relevance of the observed structures. While it would be possible to display all combinations, doing so would result in excessive and overwhelming visual output. Instead, we restrict the presentation to a focused set of examples that best highlight key features:

\begin{itemize}
    \item In \ref{sec:app_JHTDB}, we consider test case 2 and focus on the pressure field $\Pres$, which provides strong spatiotemporal contrast and helps evaluate the model’s ability to capture coherent flow features.
    
    \item In \ref{sec:app_TurbRad}, test case 0 is selected, and the density field $\density$ is used due to its capacity to highlight both large-scale gradients and small-scale variations across time.
    
    \item In \ref{sec:app_MHD}, we choose test case 0 and visualize the $B_y$ component of the magnetic field on the 2D slice at $z = 15$. Additionally, full 3D visualizations of the flow and magnetic field magnitudes, $\norm{\uv}$ and $\norm{\bft{B}}$, are provided to offer insight into the spatial structure and to demonstrate the model’s capability to generalize to 3D representations.
\end{itemize}

\subsubsection{Transonic cylinder} \label{sec:app_JHTDB}

\fref{fig:Cylinder_time_series} reveals a marked contrast between the SDE variants, particularly between the regularized and non-regularized formulations. As discussed in \sref{sec:res_JHTDB}, the regularized models preserve coherent flow structures over longer time horizons, whereas the VE SDE, in particular, rapidly degrades and loses meaningful features. This effect is especially pronounced near the cylinder, where the regularized models accurately capture shock patterns and fine-scale interactions. In contrast, the VE formulation exhibits noticeable noise and fading dynamics in both cases.

At early times, however, all models exhibit excellent visual accuracy. Shock waves are well synchronized, and sharp features are faithfully reproduced across the board. The spatial scale is globally preserved, indicating that the models effectively capture the initial dynamics. As time advances, the task becomes significantly more challenging due to increasing complexity and divergence in the flow. Despite this, the regularized formulations maintain predictive coherence, demonstrating their robustness in the long-term regime. Among all models, the regularized VP SDE consistently delivers the best performance, both in preserving structure and stabilizing long-range behavior.

\subsubsection{TurbRad} \label{sec:app_TurbRad}

\fref{fig:TurbRad_time_series} reveals less differences between SDE formulations. Both regularized and non-regularized variants manage to preserve the large-scale, membrane-like structures that characterize the flow. In terms of temporal dynamics, the evolution of chaotic patterns appears consistent with the ground truth, suggesting that the models adequately capture the rate of evolution of the simulation. The VE SDE formulation, however, stands out as the noisiest, exhibiting higher levels of spatial irregularity. This behavior aligns with the results presented in \ref{sec:app_JHTDB}, where similar instability was observed for the VE configuration. Notably, gradient-like features are still discernible on the left side of the domain, indicating partial retention of coherent structures even under noisy conditions.

Despite these observations, there is no clear advantage of regularization across all cases. Both regularized and non-regularized models exhibit comparable large-scale behavior, but some nuances emerge. The regularized sub-VP model tends to recover more fine-grained details, offering enhanced visual richness in localized regions. In contrast, the non-regularized VP variant appears smoother overall, producing a flow field that more closely resembles the reference solution in terms of structure and coherence. These differences shows the trade-offs between regularization and fidelity in turbulent flow reconstruction, with different models excelling in different qualitative aspects.

\subsubsection{MHD} \label{sec:app_MHD}

\begin{figure}[t]
    \centering
    \includegraphics[width=.9\textwidth]{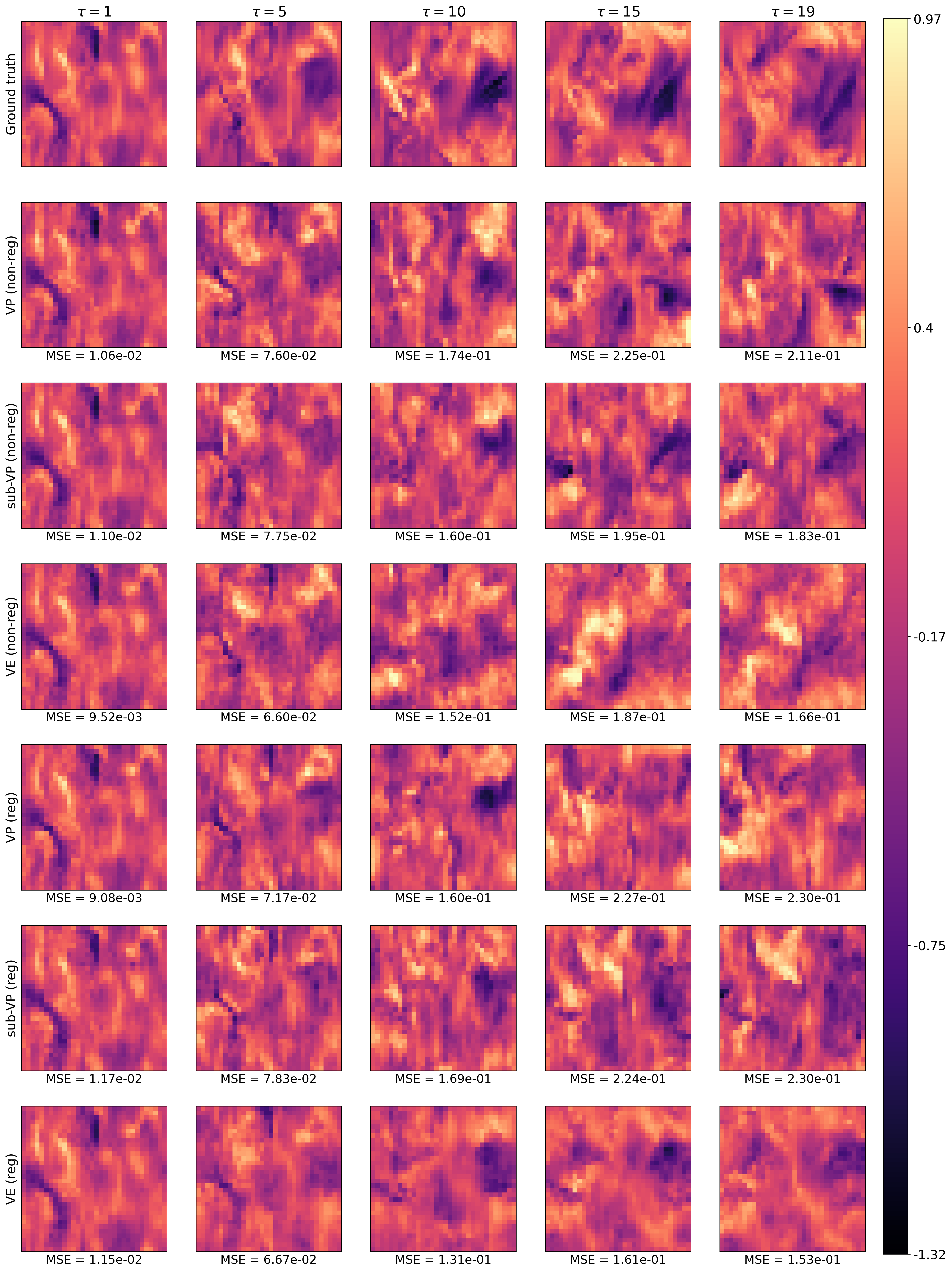}
    \caption{2D temporal visualization of the magnetic field component $B_y$ at $z = 15$ for test case 0 of the MHD dataset.}
    \label{fig:MHD_time_series}
\end{figure}

\begin{figure}[t]
    \centering
    \includegraphics[width=\textwidth]{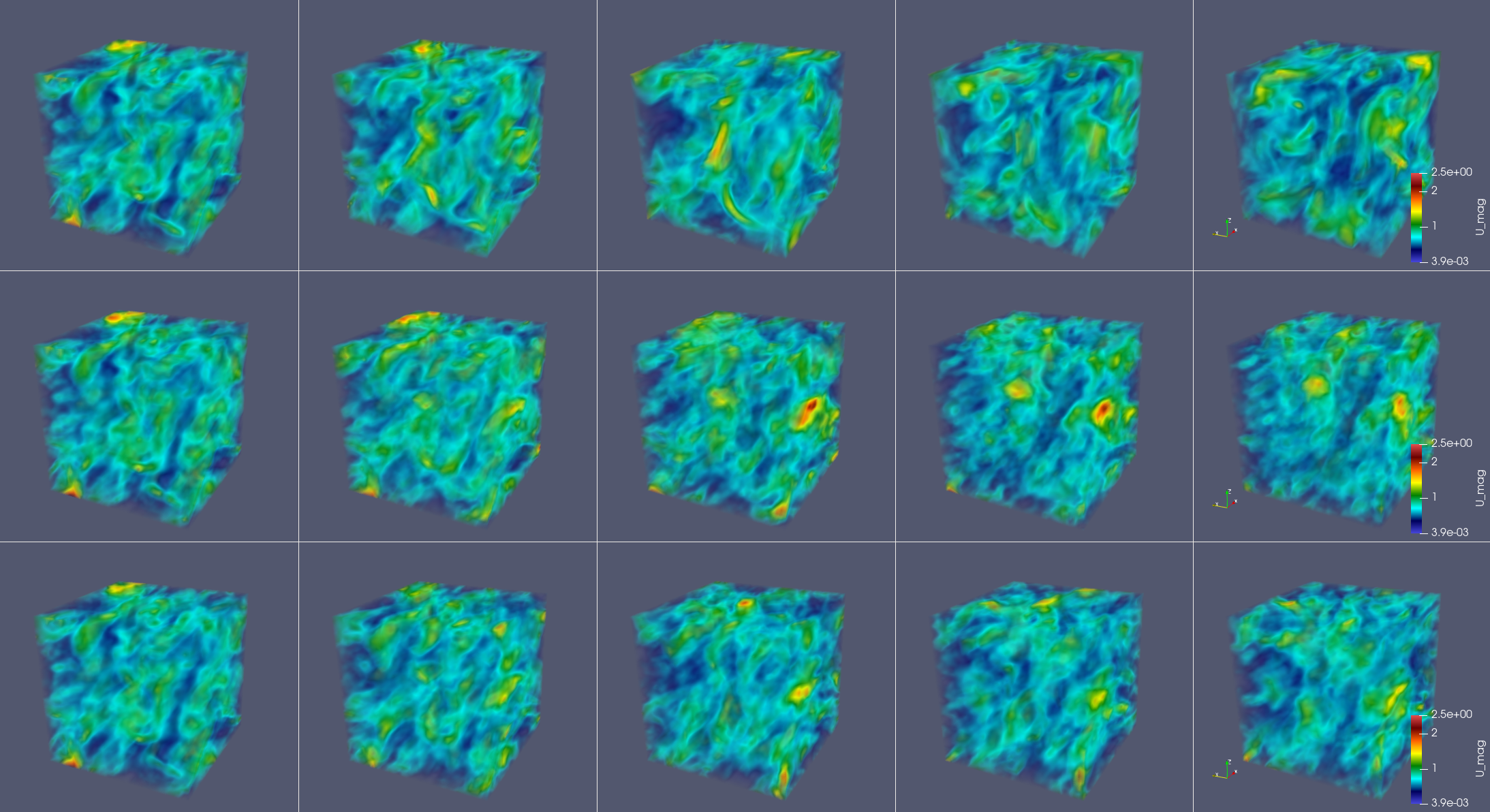}

    \vspace{-.5em}
    
    \caption{Visualization of the velocity field magnitude $\norm{\uv}$ at different time steps $\tau = 1, 5, 10, 15, 20$ (columns) for test case 0 of the MHD dataset. Each column corresponds to a specific time step. The first row displays the ground-truth data, the second row shows predictions from the non-regularized VP SDE model, and the third row shows the predictions obtained using the regularized VP SDE.}
    \label{fig:VP_U_mag}

    \vspace{1em}

    \centering
    \includegraphics[width=\textwidth]{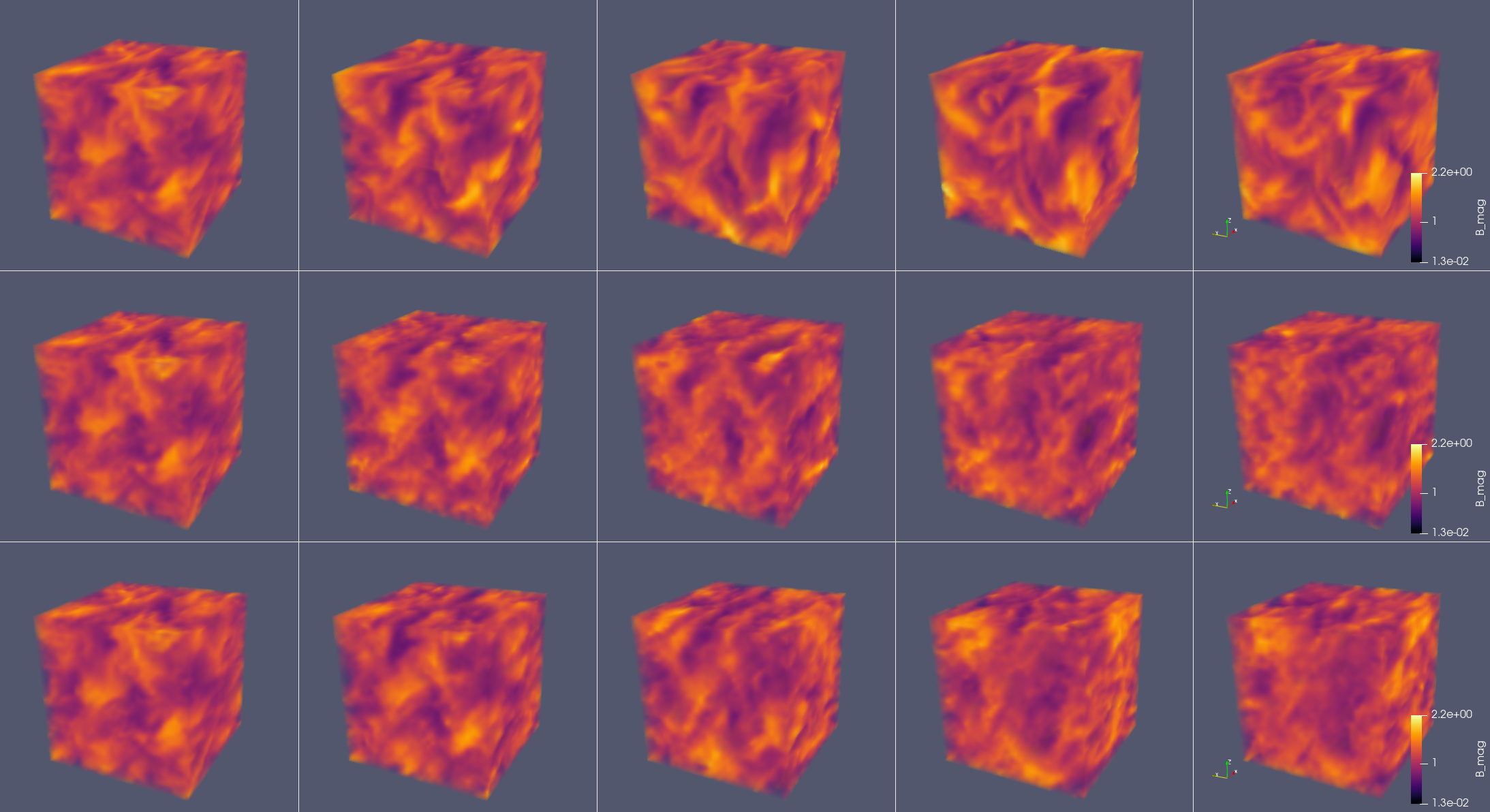}

    \vspace{-.5em}
    
    \caption{Visualization of the magnetic field magnitude $\norm{\bft{B}}$ at different time steps $\tau = 1, 5, 10, 15, 20$ (columns) for test case 0 of the MHD dataset. Each column corresponds to a specific time step. The first row displays the ground-truth data, the second row shows predictions from the non-regularized VP SDE model, and the third row shows the predictions obtained using the regularized VP SDE.}
    \label{fig:VP_B_mag}
    
\end{figure}

\fref{fig:MHD_time_series} displays a 2D temporal visualization of the magnetic field component $B_y$ at the slice $z = 15$ for test case 0 of the MHD dataset. This figure supports the findings discussed in \sref{par:mhd results}. Unlike the density field, the unregularized VE model does not exhibit evident noise in its prediction of $B_y$. All models show a clear structural divergence from the ground truth after $\tau = 5$, although the evolution remains globally coherent. However, as these are 2D slices extracted from a 3D prediction, evaluating the full spatial accuracy remains inherently limited. Nonetheless, at $\tau = 5$, certain structures exhibit noticeable geometrical coherence with the reference, indicating that the models capture some aspects of the underlying dynamics. Beyond this point, structural alignment degrades, and no model consistently demonstrates superior fidelity. The MSE remains temporally stable, suggesting a gradual and uniform increase in deviation, yet some degree of large-scale structural coherence persists across time.

\fref{fig:VP_U_mag} and \fref{fig:VP_B_mag} provide 3D renderings that offer further insight into the temporal dynamics on test case 0. While some structural coherence can be also identified around $\tau = 5$, the overall correspondence with the ground truth is less visually compelling. These visualizations primarily serve to illustrate temporal consistency rather than perfect spatial matching. Still, the predicted fields align well in terms of scale and order of magnitude, suggesting reasonable model behavior despite the complexity of the MHD dynamics.

\clearpage  






\bibliographystyle{unsrtnat} 
\bibliography{elsarticle-template-num.bbl}


\end{document}